\documentclass[acmtog,screen,nonacm]{acmart}

\usepackage{booktabs} 

\usepackage[ruled]{algorithm2e} 
\usepackage{nicefrac, xfrac}
\usepackage{adjustbox}
\usepackage[pdftex]{pict2e}

\SetAlFnt{\small}
\SetAlCapFnt{\small}
\SetAlCapNameFnt{\small}
\SetAlCapHSkip{0pt}

\acmJournal{TOG}

\copyrightyear{2023}
\acmYear{2023}
\setcopyright{acmlicensed}\acmConference[SA Conference Papers '23]{SIGGRAPH Asia 2023 Conference Papers}{December 12--15, 2023}{Sydney, NSW, Australia}
\acmBooktitle{SIGGRAPH Asia 2023 Conference Papers (SA Conference Papers '23), December 12--15, 2023, Sydney, NSW, Australia}
\acmPrice{15.00}
\acmDOI{10.1145/3610548.3618201}
\acmISBN{979-8-4007-0315-7/23/12}

\usepackage[inline]{enumitem} 
\usepackage{color}
\usepackage{multirow}
\usepackage[makeroom]{cancel}

\definecolor{turquoise}{cmyk}{0.65,0,0.1,0.3}
\definecolor{purple}{rgb}{0.65,0,0.65}
\definecolor{dark_purple}{rgb}{0.5,0,0.5}
\definecolor{dark_green}{rgb}{0, 0.5, 0}
\definecolor{orange}{rgb}{0.8, 0.6, 0.2}
\definecolor{red}{rgb}{0.8, 0.2, 0.2}
\definecolor{darkred}{rgb}{0.6, 0.1, 0.05}
\definecolor{blueish}{rgb}{0.0, 0.3, .6}
\definecolor{light_gray}{rgb}{0.7, 0.7, .7}
\definecolor{pink}{rgb}{1, 0, 1}
\definecolor{greyblue}{rgb}{0.25, 0.25, 1}

\newcommand{\ts}[1]{{\color{light_gray}#1}}

\newcommand{\ApproachName}{ShaDDR\xspace}
\begin{document}

\title[ShaDDR]{ShaDDR: Interactive Example-Based Geometry and Texture Generation via 3D Shape Detailization and Differentiable Rendering}

\author{Qimin Chen}
\affiliation{
 \institution{Simon Fraser University}
 \country{Canada}
}
\email{qca43@sfu.ca}

\author{Zhiqin Chen}
\affiliation{
 \institution{Simon Fraser University}
 \country{Canada}
}
\email{zhiqinc@sfu.ca}

\author{Hang Zhou}
\affiliation{
 \institution{Simon Fraser University}
 \country{Canada}
}
\email{haz163@sfu.ca}

\author{Hao Zhang}
\affiliation{
 \institution{Simon Fraser University}
 \country{Canada}
}
\affiliation{
 \institution{Amazon}
 \country{Canada}
}
\email{haoz@sfu.ca}

\begin{abstract}
We present ShaDDR, an {\em example-based\/} deep generative neural network which produces a high-resolution textured 3D shape through {\em geometry detailization\/} and
{\em conditional texture generation\/} applied to an input coarse voxel shape.
Trained on a small set of detailed and textured exemplar shapes, our method learns to detailize the geometry via {\em multi-resolution\/} voxel upsampling and generate textures on voxel surfaces via differentiable rendering against exemplar texture images from a few views. The generation is interactive, taking less than 1 second to produce a 3D model with voxel resolutions up to $512^3$. The generated shape preserves the overall structure of the input coarse voxel model, while the style of the generated geometric details and textures can be manipulated through learned latent codes.
In the experiments, we show that our method can generate higher-resolution shapes with plausible and improved geometric details and clean textures compared to prior works. Furthermore, we showcase the ability of our method to learn geometric details and textures from shapes reconstructed from real-world photos. In addition, we have developed an interactive modeling application to demonstrate the generalizability of our method to various user inputs and the controllability it offers, allowing users to interactively sculpt a coarse voxel shape to define the overall
structure of the detailized 3D shape.
Code and data are available at \url{https://github.com/qiminchen/ShaDDR}.

\end{abstract}



\begin{teaserfigure}
\vspace{-3mm}
  \includegraphics[width=\textwidth]{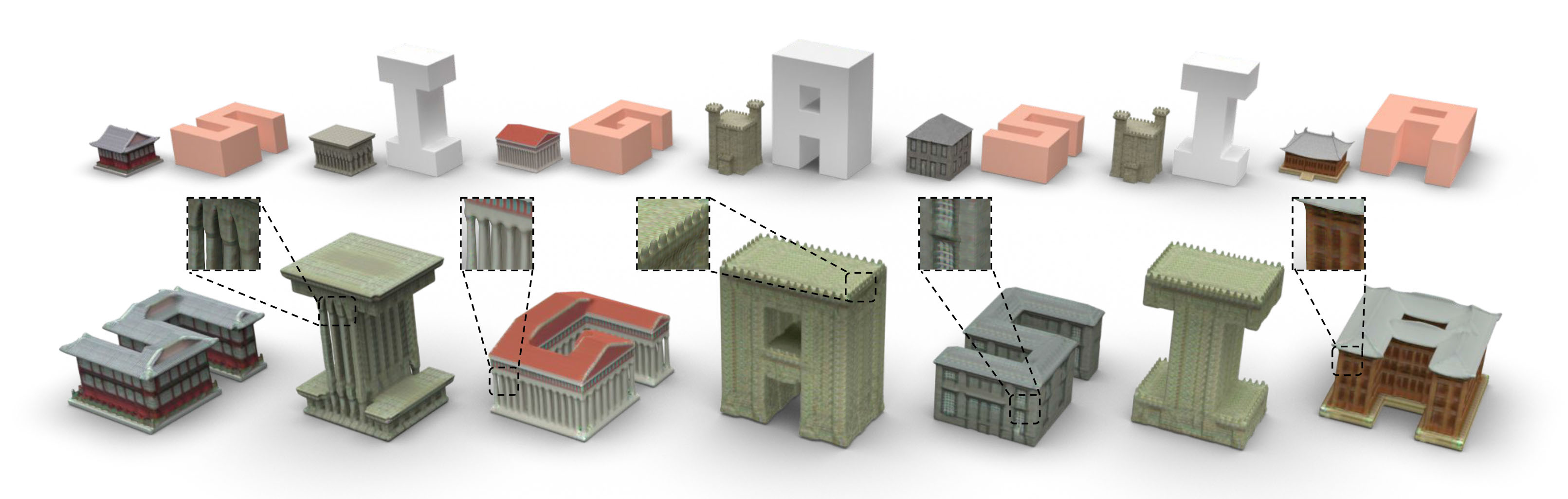}
  \vspace{-9mm}
  \caption{Given a coarse voxel shape and a textured exemplar shape (top row), our network 
  generates a {\em geometrically detailized\/} and {\em textured\/} version of the coarse shape (bottom row) in {\em less than 1 second\/}, with geometry and texture generations both conditioned on the exemplar. See zoom-ins for some details.}
  \label{fig:teaser}
\end{teaserfigure}

\maketitle

\section{Introduction}
\label{sec:intro}

Deep generative models for 3D shapes have made significant advances in recent years and lately,
the progress has been propelled by exciting developments on diffusion and large language models 
to improve usability via text prompting and generality through zero-shot learning. 
However, these new advances still do not alleviate fundamental issues such as lack of 
fine-grained control, whether the generation is from noise or texts, and slow speed, especially when diffusion
models are employed. Case in point, state-of-the-art text-prompted, diffusion-based 3D generator, Magic3D~\cite{lin2023magic3d}, currently takes about 40 minutes to produce a result. Furthermore, like 
most results of this kind, e.g., DreamFusion~\cite{poole2022dreamfusion}, Get3D~\cite{gao2022get3d}, and
Make-it-3D~\cite{tang2023makeit3d}, the generated geometry is low-resolution and lacks details.

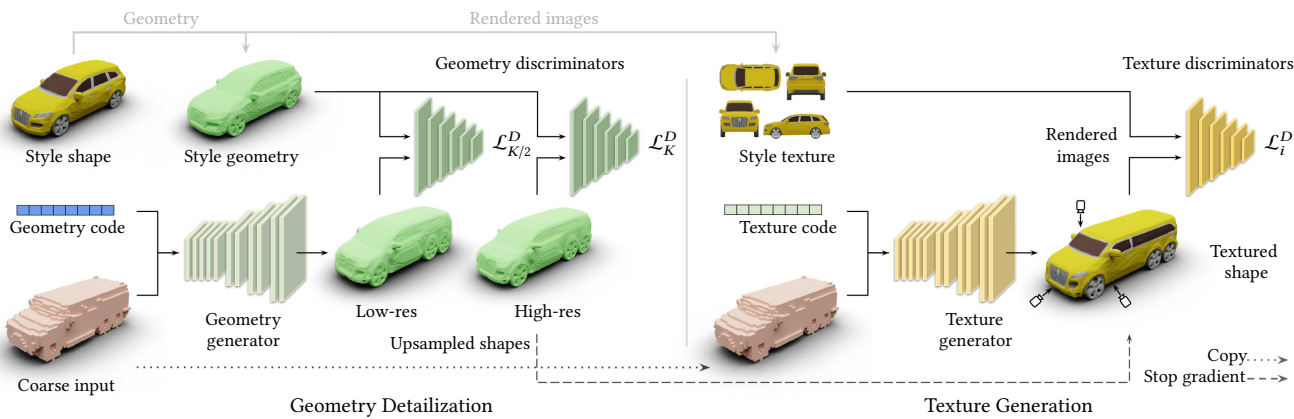
\begin{figure*}
\begin{picture}(510,157)
  \put(0,10){\includegraphics[width=0.98\linewidth]{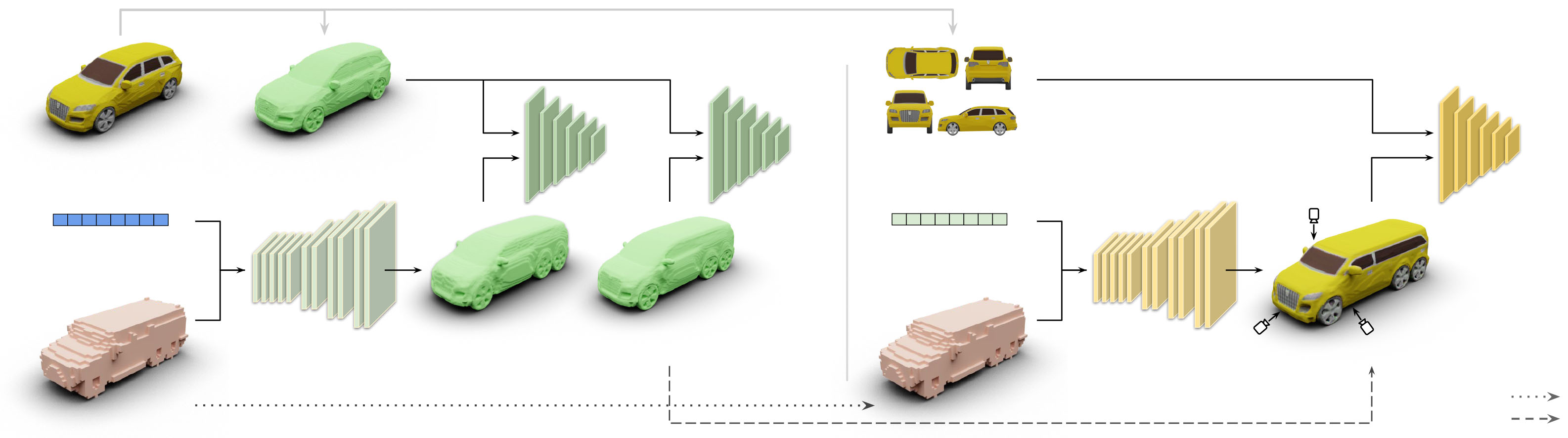}}
  
  \put(17, 11){\footnotesize Coarse input}
  \put(14, 71){\footnotesize Geometry code}
  \put(20, 98){\footnotesize Style shape}
  \put(80, 98){\footnotesize Style geometry}
  \put(57, 150){\footnotesize \ts{Geometry}}
  \put(188, 150){\footnotesize \ts{Rendered images}}
  \put(88, 36){\footnotesize Geometry}
  \put(88.5, 28){\footnotesize generator}
  \put(120, 3){\small Geometry Detailization}
  \put(205, 39){\footnotesize High-res}
  \put(145, 39){\footnotesize Low-res}
  \put(158, 27){\footnotesize Upsampled shapes}
  \put(255, 103){\footnotesize $\mathcal{L}_{K}^{D}$}
  \put(196, 103){\footnotesize $\mathcal{L}_{\nicefrac{K}{2}}^{D}$}
  \put(175, 133){\footnotesize Geometry discriminators}
  
  \put(290, 71){\footnotesize Texture code}
  \put(290, 98){\footnotesize Style texture}
  \put(368, 36){\footnotesize Texture}
  \put(365.5, 28){\footnotesize generator}
  \put(360, 3){\small Texture Generation}
  \put(406, 106){\footnotesize Rendered}
  \put(410, 98){\footnotesize images}
  \put(488, 103){\footnotesize $\mathcal{L}_{i}^{D}$}
  \put(435, 133){\footnotesize Texture discriminators}
  \put(468, 62){\footnotesize Textured}
  \put(473.5, 54){\footnotesize shape}
  
  \put(443, 14.5){\footnotesize Stop gradient}
  \put(467, 22.5){\footnotesize Copy}
\end{picture}
\vspace{-6mm}
  \caption{An overview of our \ApproachName's two-phase solution pipeline and network architecture, for which the input ``style shape'' provides the exemplars for both detailed geometry and multi-view textures. Conditioned on the geometry code, the geometry generator upsamples a coarse input voxel grid into detailed geometries in multiple (two) resolutions, $(\nicefrac{K}{2})^3$ and $K^3$. The geometry discriminators enforce the local patches of the upsampled geometries to be plausible with respect to the target geometry style. The texture generator takes in the texture code and the same coarse voxels and synthesizes 3D volumetric textures for the upsampled geometry. The generated geometry and textures are rendered into 2D images from different views, and the texture discriminators enforce the local patches of the rendered images to be plausible with respect to the target texture style.}
  \vspace{-3mm}
  \label{fig:network}
\end{figure*}
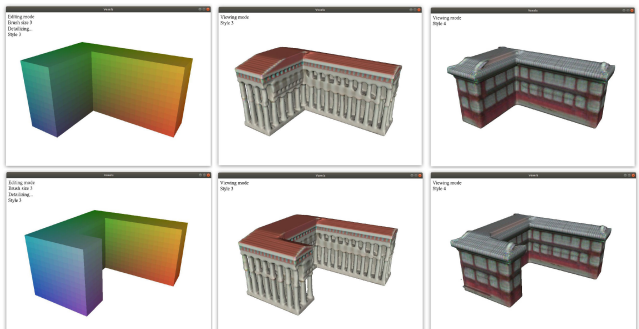
\begin{figure}
\begin{picture}(244, 142)
  \put(0, 10){\includegraphics[width=1.0\linewidth]
  {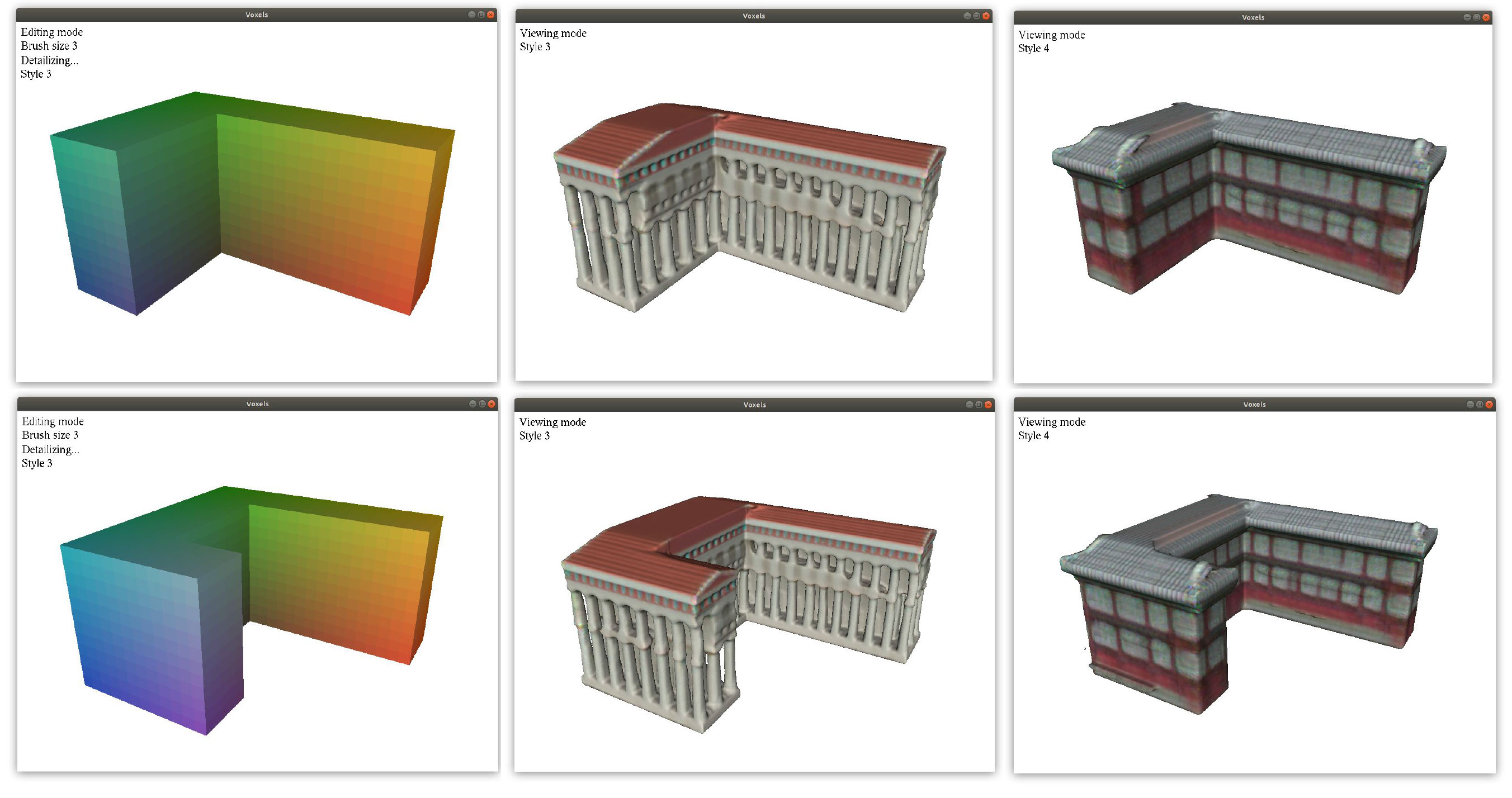}}
  \put(6, 0){\small (a) Interactive editing}
  \put(90, 0){\small (b) \ApproachName result \#1}
  \put(168, 0){\small (c) \ApproachName result \#2}
\end{picture}
  \vspace{-5mm}
  \caption{A GUI for interactive modeling by \ApproachName. Users can edit coarse voxels and visualize different detailized and textured shapes.}
  \vspace{-3mm}
  \label{fig:gui}
\end{figure}

DECOR-GAN~\cite{chen2021decor} is a recent work that takes a more traditional approach to generate higher
resolution geometries, through a process called {\em geometry detailization\/}. Specifically, their deep generative network 
{\em stylizes\/} an input coarse voxel shape, via voxel {\em upsampling\/}, conditioned on
an input exemplar shape with geometric details.
With such a modeling approach, artists retain some level of fine-grained control as they can freely edit the coarse 
voxels to define the overall structure of the detailized shape. The detailization is also fast as the network takes 
one single forward pass with both the generator and discriminator built by 3D CNNs.
However, with only a single-resolution upsampling, the generated results still cannot transfer geometric details at
finer scales. More critically, like most existing approaches for deep generative modeling of 3D shapes, DECOR-GAN only 
focuses on generating shape geometries and neglects the equally vital aspect of textures.

In this paper, we present an {\em example-based\/} generative network which produces high-resolution {\em textured\/} 
3D shapes through geometry detailization, akin to DECOR-GAN, and texture generation via differentiable
rendering. As show in Figure~\ref{fig:teaser}, our network,
coined \ApproachName~(for SHApe Detailization and Differentiable Rendering), transfers the geometric details and
textures from an exemplar textured shape which provides the geometry and texture ``styles'' or conditions for the
generative task. The generation is {\em interactive\/}, taking less than 1 second to produce a $256^3$-voxel building model.

Technically, our method operates in two separate phases, as illustrated in Figure~\ref{fig:network}. The geometry 
detailization phase is built on DECOR-GAN, sharing the same input setting and similar network architecture, except 
that we exploit the intrinsic hierarchical nature of voxel upsampling to generate detailizations at {\em multiple resolutions\/} 
(two in the current implementation). The key advantage achieved, as a direct benefit from additional supervision at 
intermediate-level voxel resolutions, is higher-quality generation of finer-scale geometric details over DECOR-GAN; 
see the several zoom-ins in Figure~\ref{fig:teaser} and a comparison presented in Figure~\ref{fig:geometry_ablation}.

The texture generator takes as input exemplar texture {\em images\/} from a small number of views (four or five depending on
the object categories), as well as the coarse 
voxel shape which was fed to the geometry generator. It synthesizes 3D volumetric textures for the highest-resolution
upsampled geometry from the first phase. The texture synthesis is learned by employing {\em image-space\/} GANs via 
differentiable rendering against the multi-view exemplar textures. In our work, the texture exemplars are multi-view 
projections from a textured 3D shape.
In general, such images may be obtained without a pre-defined 3D model; they may be sampled from  
a neural radiance field (NeRF)~\cite{mildenhall2021nerf} or taken as photographs of real-world objects (see 
the first application in Section~\ref{subsec:app}).

To our knowledge, \ApproachName~represents the first deep generative model that upsamples a coarse voxel model into 
a fully \textit{textured} and geometrically detailed 3D shape. Compared to DECOR-GAN, our method shows a clear advantage
in geometry detailization owing to its multi-resolution generator, while the conditionally generated texture significantly
improves the richness and expressiveness of the final shape. At the same time, all the merits of DECOR-GAN are retained, 
including the kind of fine-grained structural control for artists, interactive generation to facilitate exploratory 3D modeling, and 
effective reuse of existing digital assets. With the \ApproachName network, a large variety of high-resolution and detailed 3D 
geometries with textures can be easily created from a small set of exemplars, as demonstrated by extensive experiments in
Section~\ref{sec:results}.

Finally, we show two applications: a) detailed 3D shape generation with textures provided from photographs of real-world
objects; 2) an interactive modeling GUI for a user to create detailed models by sculpting coarse voxels; see Figure~\ref{fig:gui}
for a few screenshots and the supplementary video for interactive demos\footnote{Note that the model generation time within the GUI is about 2s, which is longer than that of pure network inference (<1s) due to mesh and texture exporting overhead.}.

\section{Related work}
\label{sec:rw}

Our work is closely related to 3D generative models and texture synthesis. We also discuss relevant works on few-shot generative models, i.e., models that can generate shapes when given only one or a few high-quality exemplars.

\vspace{5pt}

\emph{3D Generative Models.}
Following the introduction of variational autoencoders (VAEs) \cite{kingma2013auto}, generative adversarial networks (GANs) \cite{goodfellow2020generative}, autoregressive models \cite{van2016conditional}, and normalizing flows \cite{rezende2015variational}, various deep generative models for 3D shapes have been developed, utilizing a wide range of shape representations, including point clouds \cite{achlioptas2018learning,fan2017point,yin2019logan}, voxels \cite{choy20163d,wu2016learning,hane2017hierarchical}, deformable meshes \cite{groueix2018papier,wang2018pixel2mesh,zhangimage}, part-based structural graphs \cite{li2017grass,mo2019structurenet,gao2019sdm,gao2021tm}, and neural implicit functions \cite{chen2019imnet,mescheder2019occupancy,park2019deepsdf}.
At the same time, another line of works, including HoloGAN~\cite{Nguyen-Phuoc_2019_ICCV} and others~\cite{chan2021pi,chan2022efficient,niemeyer2021giraffe}, learn 3D-aware image synthesis.

Recent methods \cite{zeng2022lion,hui2022neural,cheng2023sdfusion} based on diffusion probabilistic models \cite{sohl2015deep,ho2020denoising} have achieved significant improvements in the geometric details of the generated shapes, yet they solely focus on geometry synthesis and do not generate textures.
Get3D \cite{gao2022get3d} relies on differentiable rendering, differentiable Marching Tetrahedra \cite{shen2021dmtet}, and 2D GANs, therefore it is able to generate textured meshes; however, it does not provide control over the generated shapes.
3DStyleNet \cite{Yin_2021_ICCV_3dstylenet} transfers the geometry and texture style from one shape to another, yet its geometry generation is achieved via deforming the source mesh, thus it cannot create contents with varying topology.
Other works \cite{poole2022dreamfusion,lin2023magic3d,melaskyriazi2023realfusion} based on differentiable rendering and test-time optimization can synthesize shape with input text or single image conditions. Since these methods overfit a NeRF \cite{mildenhall2021nerf} model for each different input, they take on the order of many minutes to hours to generate a single shape.
Besides, all these methods need to be trained with significant amounts of high-quality 3D shapes or 2D images, which are not always easily available.

In contrast, our method only needs a few (8-16) detailed and textured shapes to provide geometry and texture supervision. During test time, our method can upsample a given coarse voxel grid into a detailed shape with textures in less than 1 second. Such an interactive performance enables artists and casual users to create high-quality 3D models interactively with high controllability.

\vspace{5pt}

\emph{Few-shot Generative Models.}
Several classic methods from computer graphics are able to synthesize new content by copying local patches from a given source and joining those patches together to form the target object.
Image quilting \cite{efros2001image} is able to synthesize texture images of arbitrary sizes given an exemplar texture image. Similarly, \cite{merrell2007example,merrell2008continuous,merrell2010model} can synthesize new 3D models given an exemplar 3D shape; mesh quilting \cite{zhou2006mesh} can transfer geometric textures from a geometric texture patch to the surface of a coarse shape; and MeshMatch \cite{chen2012non} can transfer color textures from one textured shape to another textureless shape.

In the deep learning era, the explicit copy-join steps have been replaced by local patch discriminators such as PatchGAN \cite{isola2017image}. SinGAN \cite{shaham2019singan}, as well as earlier work by \cite{zhou2018non}, learns the distribution of patches within a given image in different scales, and is able to generate diverse images that carry the same content and texture as the given image.
Similarly, works by \cite{wu2022learning} and \cite{Li_2023_CVPR} learn 3D voxel patches and is able to generate diverse 3D shapes of the same style given a single exemplar shape.
3inGAN \cite{karnewar_3InGan_3dv_22} and SinGRAV \cite{wang2022singrav} are based on NeRF \cite{mildenhall2021nerf} and they can generate diverse scenes from an exemplar scene.
However, these methods perform unconditional generation and do not provide control over the generated shapes.
SketchPatch \cite{fish2020sketchpatch} aims for sketch stylization and is able to convert plain solid-lined sketches into diverse sketches with different line styles.

Most closely related to our work, DECOR-GAN \cite{chen2021decor} targets voxel detailization and is designed to generate detailed voxel shapes from input coarse voxels with respect to the style constraint controlled by a latent code. 
Inspired by pyramid-style GANs~\cite{shaham2019singan, wu2022learning}, we enable DECOR-GAN to generate higher-resolution geometric details by designing hierarchical structures in the networks. We also generate detailed textures on output shapes, making them readily usable in real applications.

\vspace{5pt}

\emph{Texture Synthesis.} 
Since classic texture synthesis methods focus on generating regular or stochastic textures on images \cite{cross1983markov,efros2001image,efros1999texture} or volumetric textures \cite{kopf2007solid}, we will mainly discuss recent works that generate semantic-aware textures on surfaces of 3D shapes via deep learning.
Various representations have been proposed to represent textures in neural networks.
Deformation-based approaches that deform a sphere primitive \cite{chen2019learning,zhang2020image,monnier2022unicorn,henderson2020leveraging,pavllo2020convolutional,li2020self,pavllo2021learning,mohammad2022clip} or a set of cuboids \cite{gao2021tm} can directly use the UV mapping defined on the primitives, therefore they only need to generate texture images.

To texture complex shapes, some works adopt texture images with pre-defined UV mapping \cite{Yin_2021_ICCV_3dstylenet,chaudhuri2021semi,TEXTure2023}, or learn the UV mapping with a neural network \cite{Chen_2022_CVPR_auvnet}.
Text2shape \cite{chen2019text2shape} adopts the voxel representation and generates RGB color for each output voxel.
Texturify \cite{siddiqui2022texturify} directly generates textures on the shape surface by introducing face convolutional operators.
The work by \cite{raj2019learning} generates multi-view images and projects the images back to the shape to obtain the textures.

Recently, with the introduction of Texture Fields \cite{Oechsle_2019_ICCV} and NeRF \cite{mildenhall2021nerf}, a number of works \cite{chan2021pi,chan2022efficient,gao2022get3d,rebain2022lolnerf,niemeyer2021giraffe,schwarz2020graf,skorokhodovepigraf,michel2022text2mesh} adopt neural fields to represent volumetric textures, possibly with view-dependent colors.
Our work is similar to Text2shape \cite{chen2019text2shape} in that we generate a grid of colored voxels. However, since our model is supervised via differentiable rendering, only surface voxels will receive gradients during training, similar to Texture Fields \cite{Oechsle_2019_ICCV}. Our method also provides control over the generated textures via a texture latent code.

\section{Method}
\label{sec:method}

In this section, we introduce our generative model, \ApproachName, that learns to upsample coarse voxels into detailed 3D geometries and corresponding textures. We represent the output geometry using an occupancy voxel grid and the output texture using a 3D RGB color grid. Specifically, given a low-resolution coarse voxel grid of $k^3$ resolution as the ``content shape'', and a pair of geometry and texture codes learned from high-resolution detailed ``style shapes'' during training, our method can generate a novel textured shape up to $K^3$ resolution that preserves the coarse structure of the content shape, while synthesizing geometric details and textures in a similar style to that of the style shape corresponding to the input geometry and texture codes. To achieve this, we devise a multi-resolution GAN operating on 3D voxels for high-resolution (up to $512^3$) geometry detailization and a set of GANs operating on 2D images rendered from the generated shapes for conditional fine texture generation.

\subsection{Multi-resolution geometry detailization}
\label{sec:multi-reso-geo-detail}

When leveraging the original DECOR-GAN \cite{chen2021decor} on super high-resolution voxels, e.g., upsampling $8$ times on each dimension, its performance degrades significantly, as shown in Figure \ref{fig:geometry_ablation} (b) where the thin structures are broken and the details are lost. This is likely due to the difficult and ambiguous nature of voxel upsampling with large upsampling factors, which can be mitigated by introducing supervision on intermediate voxel resolutions. Therefore, we employ a multi-resolution GAN by generating an intermediate lower-resolution voxel output and a final high-resolution voxel output, and applying adversarial training on both outputs.

As shown in Figure \ref{fig:network} left, given a coarse content shape represented as occupancy voxels at resolution $k^3$ and a latent geometry code representing the geometric style of a training style shape, the geometry generator is trained to upsample the content shape into two resolutions, $(\nicefrac{K}{2})^3$ and $K^3$. The two upsampled shapes are then fed into two distinct 3D CNN PatchGAN discriminators \cite{isola2017image}. We use the same auto-decoder setting as DECOR-GAN to optimize 8-dimensional learnable parameters for each style shape.

\emph{Training losses.} Following DECOR-GAN, we denote the set of detailed style shapes as $\mathcal{S}$ and the set of coarse content shapes as $\mathcal{C}$. We assume that a detailed shape $s\in \mathcal{S}$ has both geometry and texture, and the geometry represented as voxels is denoted as $s^{geo}$. Content shapes $c\in \mathcal{C}$ are coarse voxels without textures.
The geometry code representing the style of $s$ is denoted as $z_{s}^{geo}$. The geometry generator and the geometry discriminator at resolution $K^3$ are denoted as $G_{K}^{geo}$ and $D_{K}^{geo}$, respectively. Note that the discriminator is designed to be conditioned on each different style in $\mathcal{S}$ so that the style of the output shape can be controlled by $z_{s}^{geo}$.
We adopt the binary generator and discriminator masks proposed in DECOR-GAN to focus the networks' capacity on voxels close to the shape surface. The values in the masks are defined to be 1 near the surface and 0 elsewhere; see supplementary for detailed definitions. The generator masks of shapes $s$ and $c$ at resolution $K^3$ are denoted as $M_{s\cdot K}^{G}$ and $M_{c\cdot K}^{G}$, respectively. Similarly, the discriminator masks are denoted as $M_{s\cdot K}^{D}$ and $M_{c\cdot K}^{D}$. We adapt the adversarial loss from DECOR-GAN to our multi-resolution setting and only define discriminator and generator losses on the final resolution of $K^3$ for simplicity; losses for $(\nicefrac{K}{2})^3$ resolution can be derived by changing $K$ to $\nicefrac{K}{2}$. The discriminator loss is defined as:
\begin{align}
    \mathcal{L}_{K}^{D} &= \underset{s\sim \mathcal{S}}{\mathbb{E}}\frac{||(D_{K}^{geo}(s^{geo})-1)\circ M_{s\cdot K}^{D}||_{2}^{2}}{||M_{s\cdot K}^{D}||_{1}} + \mathop{\underset{s\sim \mathcal{S}}{\mathbb{E}}}_{c\sim \mathcal{C}}\frac{||D_{K}^{geo}(c_{s\cdot K}^{geo})\circ M_{c\cdot K}^{D}||_{2}^{2}}{||M_{c\cdot K}^{D}||_{1}}\;, \nonumber
\end{align}
\begin{equation}
\label{eqn:geo_dis_loss}
    c_{s\cdot K}^{geo} = G_{K}^{geo}(c,\, z_{s}^{geo})\circ M_{c\cdot K}^{G},
\end{equation}
where $\circ$ denotes element-wise multiplication, and $c_{s\cdot K}^{geo}$ is the upsampled shape of resolution $K^3$ from input coarse shape $c$ with the style of $s$.
The generator loss is defined as:
\begin{equation}
\label{eqn:geo_gen_loss}
    \mathcal{L}_{K}^{G} = \mathop{\underset{s\sim \mathcal{S}}{\mathbb{E}}}_{c\sim \mathcal{C}}\frac{||(D_{K}^{geo}(c_{s\cdot K}^{geo})-1)\circ M_{c\cdot K}^{D}||_{2}^{2}}{||M_{c\cdot K}^{D}||_{1}}.
\end{equation}

Additionally, we adapt the reconstruction loss from DECOR-GAN to the multi-resolution setting: if both the input coarse shape and the geometry code stem from the same detailed style shape, we expect the outputs of the geometry generator to be the ground truth geometry at both the intermediate and the final resolutions.
\begin{equation}
\label{eqn:geo_recon_loss}
    \mathcal{L}_{K}^{recon} = \underset{s\sim \mathcal{S}}{\mathbb{E}}\frac{||G_{K}^{geo}(s^{geo}_{\downarrow},\, z_{s}^{geo})\circ M_{s\cdot K}^{G} - s^{geo}||_{2}^{2}}{|s^{geo}|},
\end{equation}
where $|s^{geo}|$ is the volume (height $\times$ width $\times$ depth) of the voxels in $s^{geo}$, and $s^{geo}_{\downarrow}$ is the coarse content shape downsampled from $s^{geo}$.

The overall loss for geometry detailization is the sum of the GAN loss and the reconstruction loss at each resolution:
\begin{equation}
\label{eqn:geo_loss}
    \mathcal{L}_{geo} = \mathcal{L}_{K}^{G} + \mathcal{L}_{\nicefrac{K}{2}}^{G} + \mathcal{L}_{K}^{recon} + \mathcal{L}_{\nicefrac{K}{2}}^{recon}.
\end{equation}

\subsection{Texture generation via differentiable rendering}
\label{sec:rendering-based-texture-generation}

Similar to our geometry upsampling pipeline, a na\"ive way to synthesize textures on upsampled shapes is to define them as volumetric textures and directly apply 3D GANs on a grid of RGB values. However, as discussed in our ablation study in Section \ref{sec:ablation_study}, this design does not fully utilize the discriminator's capacity, as the discriminator has to account for colors of voxels in the ambient space, which are ill-defined.
Indeed, textures are 2D in nature, and only voxels on the surfaces of the shapes have well-defined colors.
Therefore, given the upsampled geometry, we first apply differentiable rendering to project the colors of the surface voxels onto a 2D image, and then employ 2D GANs for learning texture synthesis.

\emph{Differentiable rendering.}
As shown in Figure \ref{fig:network} (right), a coarse content shape at resolution $k^3$ and a latent texture code representing the texture style of a training shape are given as input to the texture generator. The generator synthesizes a $K^3$ voxel grid of RGB color values.
To render a 2D texture image, for each pixel, we shoot a ray from the posed camera and compute the first occupied voxel intersected in the upsampled geometry voxel grid. We then use the coordinates of that voxel to retrieve the corresponding RGB value from the generated voxel color grid to represent the pixel color.

\emph{Training losses.}
We denote the texture code representing the style of $s$ as $z_{s}^{tex}$. The color grid of $s$ is denoted as $s^{tex}$. The texture generator is denoted by $G^{tex}$ and the texture discriminator of view $i$ is denoted by $D_{i}^{tex}$. We also reuse the notations defined in Section \ref{sec:multi-reso-geo-detail}. The binary discriminator masks of shape $s$ and $c$ from view $i$ are denoted as $M_{s\cdot i}^{D}$ and $M_{c\cdot i}^{D}$, respectively. Similar to the geometry masks, these texture masks are used to focus the networks' capacities on non-empty regions in the rendered images; see supplementary for detailed definitions. The discriminator loss is defined as:
\begin{align}
    \mathcal{L}_{i}^{D} &= \underset{s\sim \mathcal{S}}{\mathbb{E}}\frac{||(D_{i}^{tex}(R_{i}(s^{geo},\, s^{tex}))-1)\circ M_{s\cdot i}^{D}||_{2}^{2}}{||M_{s\cdot i}^{D}||_{1}} \nonumber \\ & + \mathop{\underset{s\sim \mathcal{S}}{\mathbb{E}}}_{c\sim \mathcal{C}}\frac{||D_{i}^{tex}(R_{i}(c_{s\cdot K}^{geo},\, c_{s\cdot K}^{tex}))\circ M_{c\cdot i}^{D}||_{2}^{2}}{||M_{c\cdot i}^{D}||_{1}}\;, \nonumber
\end{align}
\begin{equation}
\label{eqn:tex_dis_loss}
    c_{s\cdot K}^{tex} = G^{tex}(c,\, z_{s}^{tex}),
\end{equation}
where $c_{s\cdot K}^{tex}$ is the synthesized color grid, and $R_{i}(\cdot, \cdot)$ is the rendering function at view $i$ given the geometry voxels and the color grid.
The generator loss is defined as:
\begin{equation}
\label{eqn:tex_gen_loss}
    \mathcal{L}_{i}^{G} = \mathop{\underset{s\sim \mathcal{S}}{\mathbb{E}}}_{c\sim \mathcal{C}}\frac{||(D_{i}^{tex}(R_{i}(c_{s\cdot K}^{geo},\, c_{s\cdot K}^{tex}))-1)\circ M_{c\cdot i}^{D}||_{2}^{2}}{||M_{c\cdot i}^{D}||_{1}}.
\end{equation}

For the reconstruction loss, we expect the rendered images of each view to be the ground truth images if the input coarse shape and the texture code stem from the same detailed style shape:
\begin{equation}
    \mathcal{L}_{i}^{recon} = \underset{s\sim \mathcal{S}}{\mathbb{E}}\frac{||R_{i}(s_{s\cdot K}^{geo},\, s_{s\cdot K}^{tex})-R_{i}(s^{geo},\, s^{tex})||_{2}^{2}}{|R_{i}(s^{geo},\, s^{tex})|}, \nonumber
\end{equation}
\begin{equation}
\label{eqn:tex_recon_loss}
    s_{s\cdot K}^{geo} = G_{K}^{geo}(s^{geo}_{\downarrow},\, z_{s}^{geo}), \;\;
    s_{s\cdot K}^{tex} = G^{tex}(s^{geo}_{\downarrow},\, z_{s}^{tex}),
\end{equation}
where $|R_{i}(s^{geo},\, s^{tex})|$ is the area (height $\times$ width) of the rendered image. $s_{s\cdot K}^{geo}$ and $s_{s\cdot K}^{tex}$ are the upsampled geometry and synthesized texture from downsampled $s^{geo}$, respectively.

The overall loss for texture generation is the sum of the GAN loss and the reconstruction loss at each view.
\begin{equation}
\label{eqn:tex_loss}
    \mathcal{L}_{tex} = \sum_{i}(\mathcal{L}_{i}^{G} + \mathcal{L}_{i}^{recon}).
\end{equation}

\subsection{Implementation details}

Our generators are designed to upsample the geometry $8$ times, i.e., $\nicefrac{K}{k}=8$.
In our experiments, we train individual models for different shape categories. For each category, we train the geometry generator first, and then train the texture generator while fixing the weights of the geometry generator.
We assume the training shapes from some categories (\textit{car}, \textit{airplane}, and \textit{chair}) are bilaterally symmetrical, therefore only generating half of the shape. We do not make symmetrical assumption for the \textit{building} and \textit{plant} categories, thus generating the whole shape.
Depending on the category, training each model for geometry detailization and texture generation on a single NVIDIA RTX 3090 Ti GPU takes 12-24 hours and 18-36 hours, respectively. The network inference takes less than 1 second per content per style. The network architectures and training details can be found in the supplementary material.

\subsection{Data preparation}
\label{sec:data_preparation}

We follow DECOR-GAN \cite{chen2021decor} to prepare occupancy voxels of $512^3$ resolution for each shape, which can be downsampled to $K^3$ and $k^3$ resolutions as style shapes and content shapes for training.
For detailed style shapes, we apply a Gaussian filter with $\sigma=1.0$ on the geometry voxels to encourage continuous optimization in GANs. 
To generate colored voxels for detailed style shapes, we first render the shapes into images from different views, and project the pixel colors in the rendered images back to the surface voxels. We inpaint the colors of non-surface voxels using nearest neighbor to obtain volumetric textures for our ablation study.

\section{Results and evaluation}
\label{sec:results}

We conduct a series of experiments on different categories to demonstrate the effectiveness and generalizability of our method.

\emph{Dataset}. We test our method on cars, airplanes, and chairs from ShapeNet \cite{chang2015shapenet}, and buildings from Houses3K \cite{peralta2020next}.
We use 3,141 cars, 1,743 airplanes, and 2,824 chairs as coarse content shapes, as in DECOR-GAN. We augment the $600$ buildings from Houses3K with $90^{\circ}$ rotations into 2,400 shapes as content shapes. We divide the content shapes into a $80\% / 20\%$ train-test split. We select $8$ style shapes for the building category and $16$ style shapes for the other categories with varying topologies and textures. We additionally test on the plant category with 300 content shapes and 20 style shapes. We use $64^3$ coarse content shapes for cars and airplanes, and $32^3$ for chairs, buildings, and plants to further remove topological details. We extract surfaces via Marching Cubes \cite{marchingcube1987}, and determine the vertex colors of the output mesh by querying the synthesized color grid.

\emph{Metrics}. We employ Intersection over Union (IoU), F-score, and Cls-score from DECOR-GAN to evaluate the quality of the generated geometry. To quantitatively measure the quality of the textures, we render images of the generated shapes and the style shapes from different views and evaluate with Fr\'echet Inception Distance (FID) \cite{heusel2017gans} and Learned Perceptual Image Patch Similarity (LPIPS) \cite{zhang2018perceptual} metrics. We use FID-all to measure the similarity between the generated textures and all ground truth textures used in our training, and FID-style to measure the similarity between the generated textures and the ground truth texture that provides the style of the generated textures. Similar to FID-style, we use LPIPS-style to evaluate the local patch similarity between generated textures and the ground truth textures of each style. The full details can be found in the supplementary material.

\subsection{Geometry detailization and texture generation}
\label{sec:result-visual}

The visual results of the car, airplane, building, chair and plant  categories can be found in Figures 
\ref{fig:car_results}-\ref{fig:plant_results},
with more results in the supplementary. Our method is able to not only produce high-resolution upsampled geometry with fine local details, e.g., the car wheels, but also synthesize high-quality texture colors, e.g., the stripes pattern of the car and the windows of the building.

\subsection{Ablation study}
\label{sec:ablation_study}

\begin{figure}
\begin{picture}(244, 160)
  \put(0, 0){\includegraphics[width=0.99\linewidth]{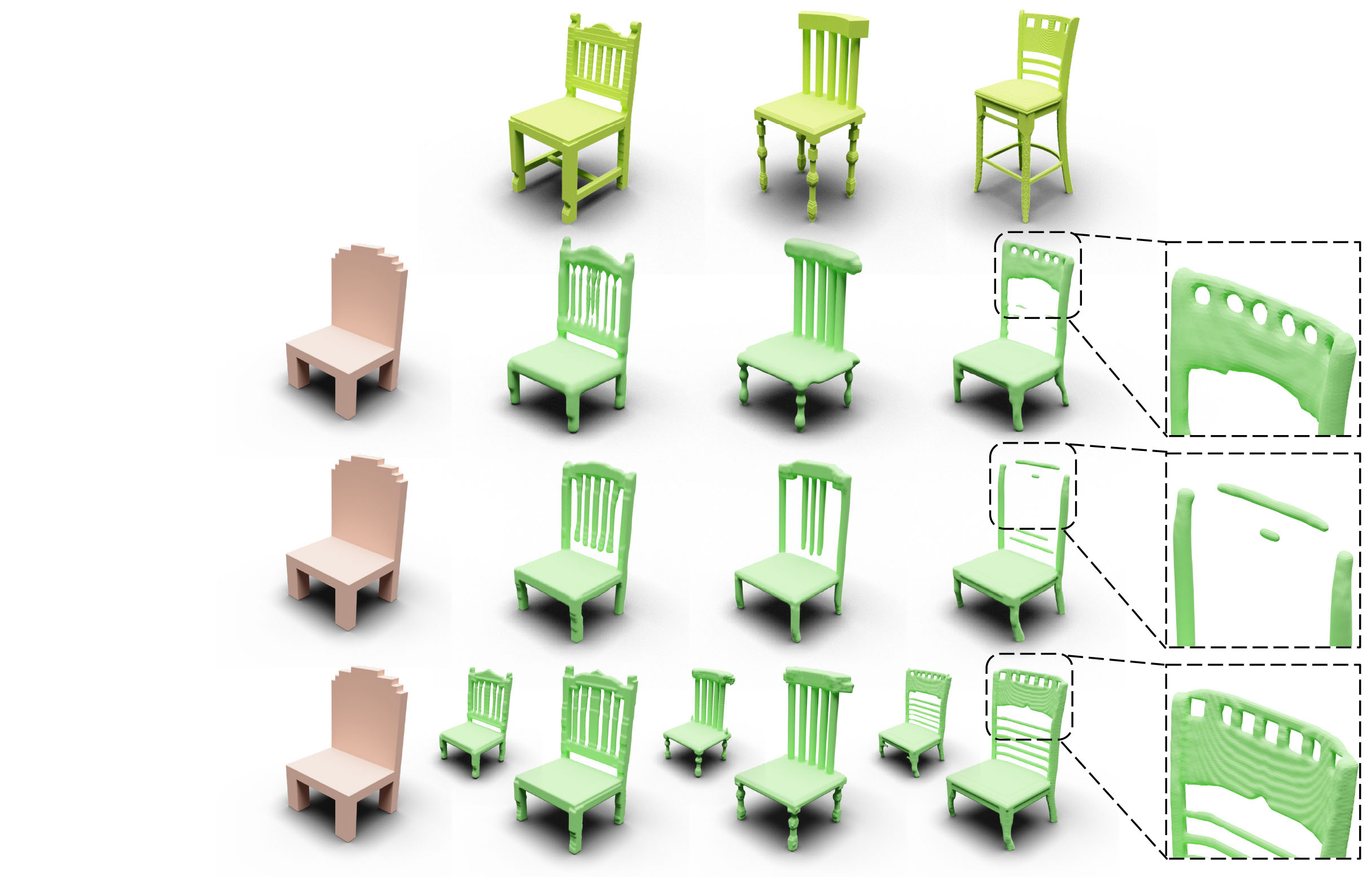}}
  \put(0, 30){\small (c) Ours}
  \put(10, 20){\small (multi-res)}
  
  \put(0, 62){\small (b) DECOR-}
  \put(11, 52){\small GAN-up}
  \put(0, 100){\small (a) DECOR-}
  \put(11, 90){\small GAN}
  \linethickness{0.005mm}
  \put(25,145){\line(36,-13){43}}
  \put(25, 127){\small Content}
  \put(44, 145){\small Geometry}
  \put(52, 137){\small Code}
  \end{picture}
  \vspace{-5mm}
  \caption{Comparison of geometry detailization between DECOR-GAN ($K=128$), DECOR-GAN-up ($K=256$), and our multi-resolution in \ApproachName~($K=256$, small figures show the intermediate voxels at $128$ resolution).}
  \label{fig:geometry_ablation}
\end{figure}
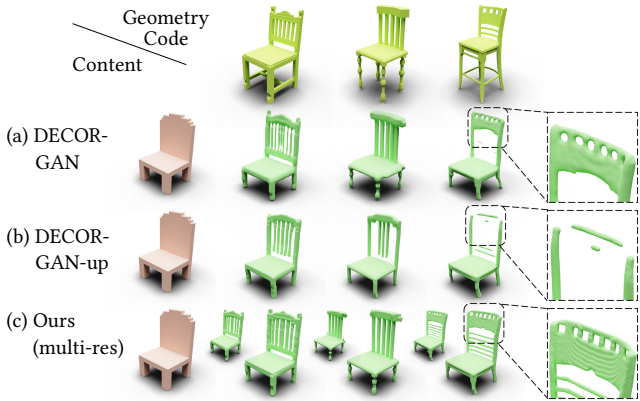

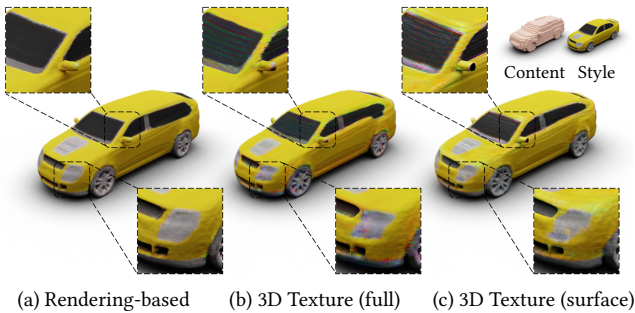
\begin{figure}
\begin{picture}(244, 110)
  \put(0, 10){\includegraphics[width=0.99\linewidth]{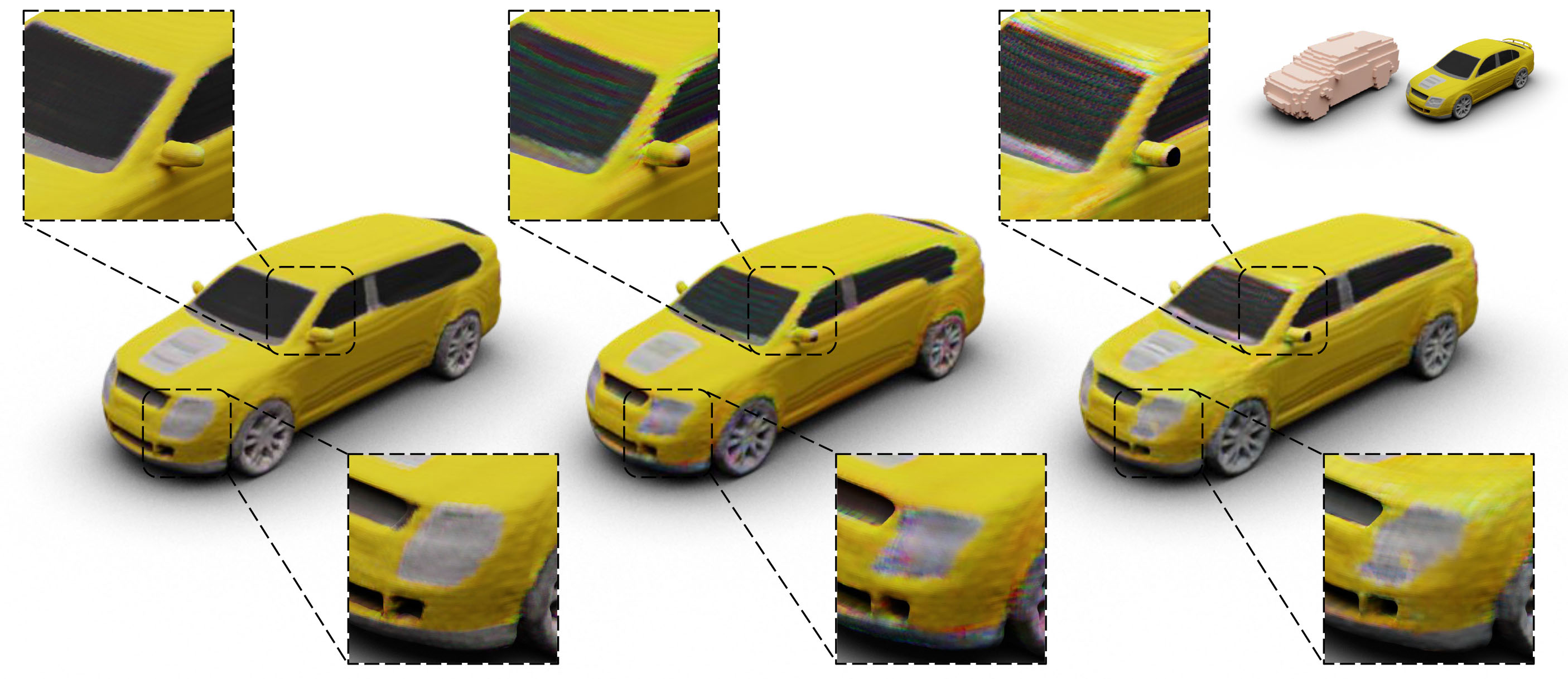}}
  \put(8, 0){\small (a) Rendering-based}
  \put(88, 0){\small (b) 3D Texture (full)}
  \put(165, 0){\small (c) 3D Texture (surface)}
  \put(192, 86){\footnotesize Content}
  \put(220, 86){\footnotesize Style}
\end{picture}
  \vspace{-5mm}
  \caption{Comparison of texture generation results between our rendering-based approach and two baselines. Please zoom in to observe the details.}
  \vspace{-3mm}
  \label{fig:texture_ablation}
\end{figure}

\begin{table}[t]
    \begin{center}
    \caption{Quantitative comparison between our multi-resolution geometry upsampling and two baselines on the chair category.}
    \label{tab:geometry_ablation}
    \vspace{-3mm}
    
    \begin{adjustbox}{width=0.475\textwidth}
    \begin{tabular}{rccccccc}
    \toprule
        & Strict- & Loose- & LP- & LP-F- & Div- & Div-F- & Cls- \\
        & IOU $\uparrow$ & IOU $\uparrow$ & IOU $\uparrow$ & score $\uparrow$ & IoU $\uparrow$ & score $\uparrow$ & score $\downarrow$ \\
    \midrule
       DECOR-GAN & 0.579 & 0.751 & 0.415 & 0.960 & 0.963 & 0.944 & 0.657 \\
       DECOR-GAN-up & \textbf{0.639} & 0.769 & 0.660 & 0.980 & 0.975 & 0.900 & 0.695 \\
       Ours (multi-res) & 0.629 & \textbf{0.771} & \textbf{0.692} & \textbf{0.985} & \textbf{1.000} & \textbf{0.969} & \textbf{0.597} \\
    \bottomrule
    \end{tabular}
    \end{adjustbox}
    \end{center}
    \vspace{-3mm}
\end{table}

\begin{table*}[t]
    \begin{center}
    \caption{Quantitative evaluation of different texture generation approaches. Lower FID and LPIPS indicate higher quality of the generated textures.}
    \label{tab:2d_vs_3d_texture}
    \vspace{-3mm}
    
    \begin{adjustbox}{width=0.98\textwidth}
    \begin{tabular}{rcccccccccccc}
    \toprule
        & \multicolumn{4}{c}{FID-all $\downarrow$} & \multicolumn{4}{c}{FID-style $\downarrow$} & \multicolumn{4}{c}{LPIPS-style $\downarrow$} \\ \cmidrule(lr){2-5} \cmidrule(lr){6-9} \cmidrule(lr){10-13} & Car & Airplane & Chair & Building & Car & Airplane & Chair & Building & Car & Airplane & Chair & Building \\
    \midrule
       3D texture (full) & 51.094 & 19.438 & 46.434 & 127.834 & 104.384 & 58.199 & 106.707 & 192.059 & 0.121 & 0.111 & 0.274 & 0.396 \\
       3D texture (surface) & 51.519 & 23.438 & \textbf{45.727} & 128.227 & 105.278 & 61.410 & 104.850 & 194.275 & 0.115 & 0.113 & 0.276 & 0.398 \\
       Rendering-based & \textbf{41.573} & \textbf{13.726} & 45.962 & \textbf{124.618} & \textbf{88.325} & \textbf{43.418} & \textbf{104.735} & \textbf{189.071} & \textbf{0.104} & \textbf{0.101} & \textbf{0.270} & \textbf{0.386} \\
    \bottomrule
    \end{tabular}
    \end{adjustbox}
    \end{center}
    \vspace{-1.5mm}
\end{table*}

\emph{Geometry detailization w/o vs. w/ multi-resolution setting.} We compare our multi-resolution \ApproachName with the original DECOR-GAN \cite{chen2021decor}, which generates single-resolution geometries. We lift the DECOR-GAN upsampling resolution, denoted as DECOR-GAN-up, to match ours for a fair comparison. Note that our multi-resolution setting has nearly the same network architectures as DECOR-GAN-up except that ours has one more layer for generating geometry at an {\em intermediate\/} resolution. 

Table \ref{tab:geometry_ablation} shows the quantitative comparsion between geometries generated by DECOR-GAN, DECOR-GAN-up, and ours on the chair category. More results can be found in the supplementary. Our method outperforms the two baselines by a reasonable margin over most of the metrics. We show a qualitative comparison in Figure \ref{fig:geometry_ablation}, where one can note that the original DECOR-GAN \cite{chen2021decor} 
is unable to provide sufficient local details, e.g., sharp edges shown in Figure \ref{fig:geometry_ablation} (a). DECOR-GAN-up directly lifts the upsampling rate twice, thus the geometry generator has to learn a more complicated upsampling space where the upsampled geometry might be locally distinctive but globally implausible. As a result, it fails to generate complete geometric structures even though the local structures may be able to follow the style shapes, but only partially, as shown in Figure \ref{fig:geometry_ablation} (b). With our multi-resolution setting, both global structures and local details of the upsampled geometry are already better than DECOR-GAN and DECOR-GAN-up at the intermediate resolution, while detailed geometric features, e.g., sharp edges and smooth surfaces, are refined at the final resolution; see Figure \ref{fig:geometry_ablation} (c). More qualitative comparisons can be found in the supplementary.

\emph{Learning textures via 3D vs. 2D supervision}. Since we generate a 3D grid of colors to represent textures, it is possible to treat the color grid as volumetric texture and supervise the learning with a 3D texture discriminator, similar to our 3D geometry discriminator. We denote this setting as \textit{3D texture (full)}. It is worth noting that texture is a surface property and learning colors of voxels inside or outside the shape is intrinsically ill-posed. Therefore, we modify the 3D texture discriminator to only learn the voxel colors near the geometry surface by applying texture discriminator masks whose values are $1$ near the surface and $0$ elsewhere, similar to $M_{s\cdot K}^{D}$ and $M_{c\cdot K}^{D}$. We denote this setting as \textit{3D texture (surface)}. For a fair comparison, we use the same generator architecture for all settings and only replace the 2D convolutions in our texture discriminator with 3D convolutions to be used in those baselines.

A qualitative comparison between our rendering-based approach and the two baselines is shown in Figure \ref{fig:texture_ablation}. Note that even though both 3D texture (full) and 3D texture (surface) can generate reasonable textures, for the local regions, e.g. windshield, window frame, and headlight, textures synthesized by our rendering-based method are cleaner and sharper compared to the two baselines whose generated textures are stained and uneven. We also provide a quantitative comparison in Table \ref{tab:2d_vs_3d_texture}. Rendering-based approach outperforms the two baselines in all metrics except FID-all in chair category. 

\subsection{Application}
\label{subsec:app}

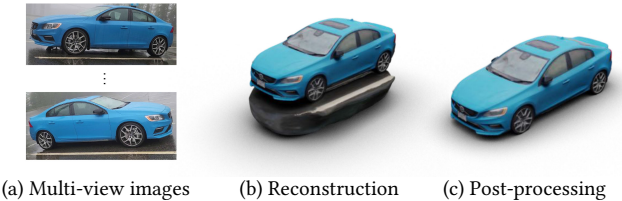
\begin{figure}
\begin{picture}(244, 75)
  \put(0, 10){\includegraphics[width=1.0\linewidth]
  {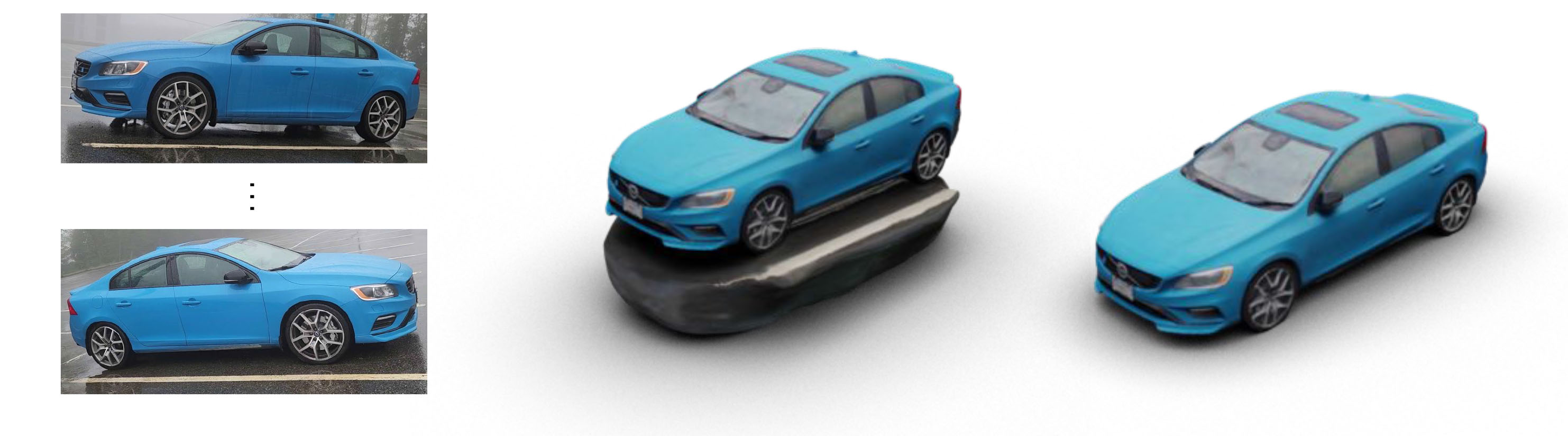}}
  \put(0, 3){\small (a) Multi-view images}
  \put(90, 3){\small (b) Reconstruction}
  \put(167, 3){\small (c) Post-processing}
\end{picture}
  \vspace{-6mm}
  \caption{We take multiple photos of a real car and reconstruct a textured mesh using NeuS \cite{wang2021neus} with post-processing.}
  \vspace{-4mm}
  \label{fig:real_data_processing}
\end{figure}

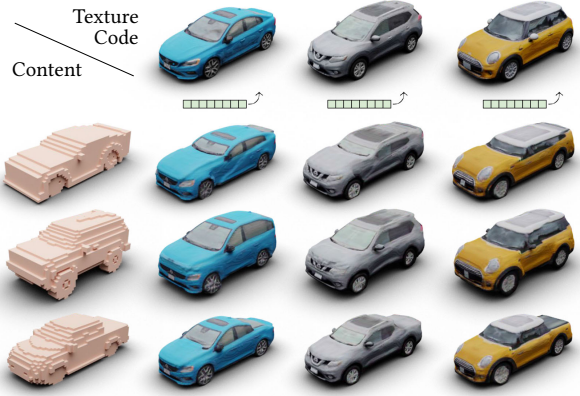
\begin{figure}
\begin{picture}(244,158)
  \put(0, 0){\includegraphics[width=0.99\linewidth]{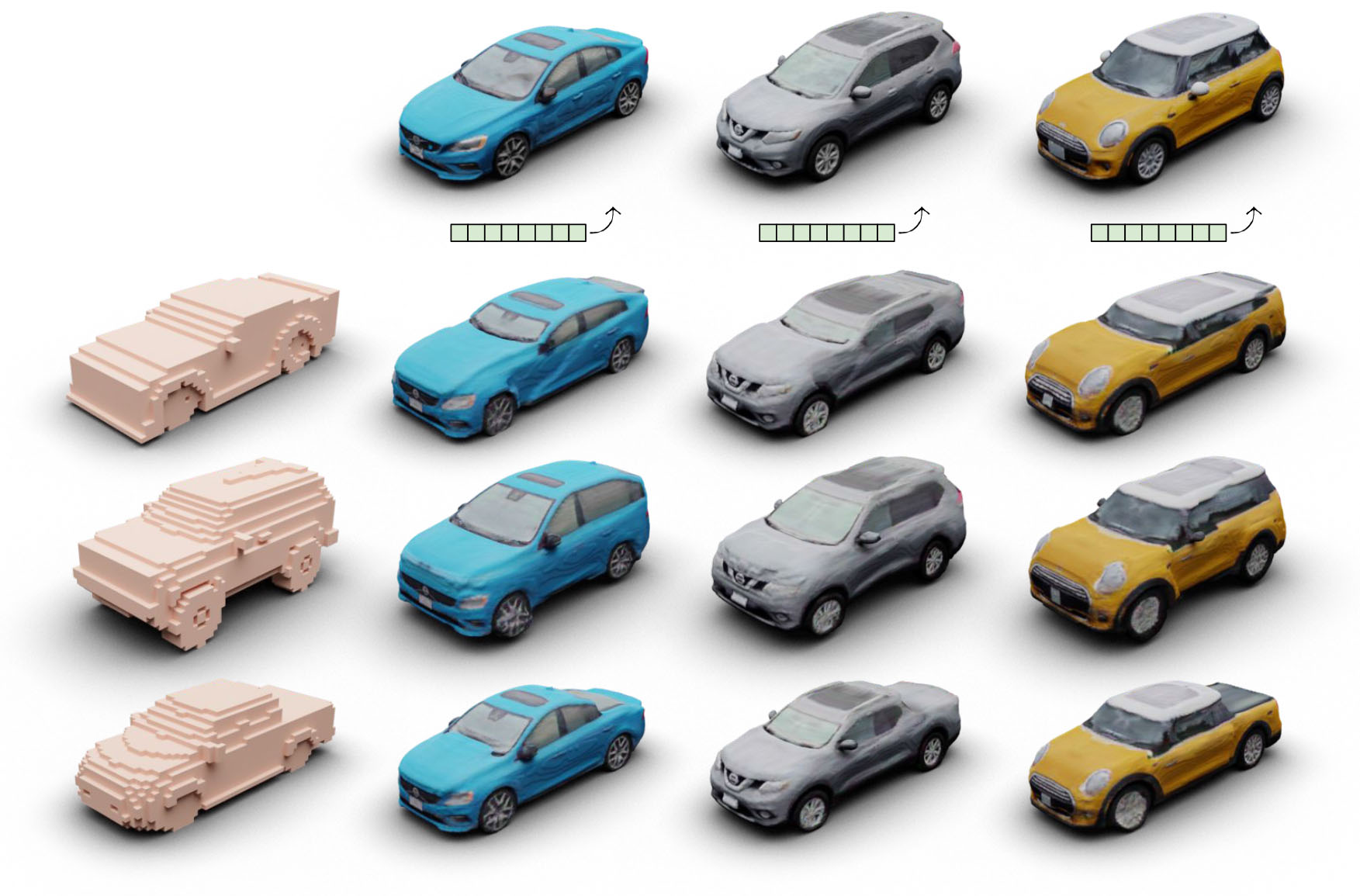}}
  \linethickness{0.005mm}
  \put(16,148){\line(18,-8){45}}
  \put(15, 128){\small Content}
  \put(38, 148){\small Texture}
  \put(46, 140){\small Code}
\end{picture}
  \vspace{-9mm}
  
  \caption{We show the input coarse content voxels on the left and the detailed style shapes with textures reconstructed from real-world photos on top. The coarse content voxels are $64^3$ and the generated shapes are $512^3$.}
  \vspace{-3mm}
  \label{fig:real_car_results}
\end{figure}

\emph{Styles from real photos}. We show that we can use the detailed shapes reconstructed from real-world images as style shapes to \emph{guide} both geometry and texture generation on new coarse content voxel shapes. We capture multi-view images from real cars in the wild, as shown in Figure \ref{fig:real_data_processing} (a). Then we use NeuS \cite{wang2021neus} to reconstruct both the geometry and the texture, as shown in Figure \ref{fig:real_data_processing} (b). Finally, we post-process the reconstructed mesh by manually removing the ground and closing holes, as shown in Figure \ref{fig:real_data_processing} (c). We then follow Section \ref{sec:data_preparation} to prepare the style shapes for training. Figure \ref{fig:real_car_results} shows the detailization results.

\emph{Generalizability and interactive editing}. We show that our method is capable of handling various coarse input shapes that are manually designed by users without further training or fine-tuning. We manually design several coarse content shapes, e.g., \textit{SIGASIA} in Figure \ref{fig:teaser}, and directly test our model trained in prior experiments in Section \ref{sec:result-visual}. Thanks to the inductive bias of 3D convolutions, i.e., convolutional operations are local and translation equivariant, our texture generator has strong generalizability to new shapes and is able to handle areas that are fully invisible in rendered views during training. Moreover, our model only needs a single forward pass to detailize a new shape, which takes less than a second, thus allowing our method to be integrated into interactive modeling tools. We have created an interactive modeling interface where users can edit a coarse content voxel, choose a style, and visualize the detailed textured shape in real time. Figure \ref{fig:gui} shows the interactive editing process and the detailization results. We also provide a video recording to show the entire editing process in the supplementary.
\section{Conclusions}

\begin{figure}
\begin{picture}(244, 185)
  \put(0, 10){\includegraphics[width=0.99\linewidth]
  {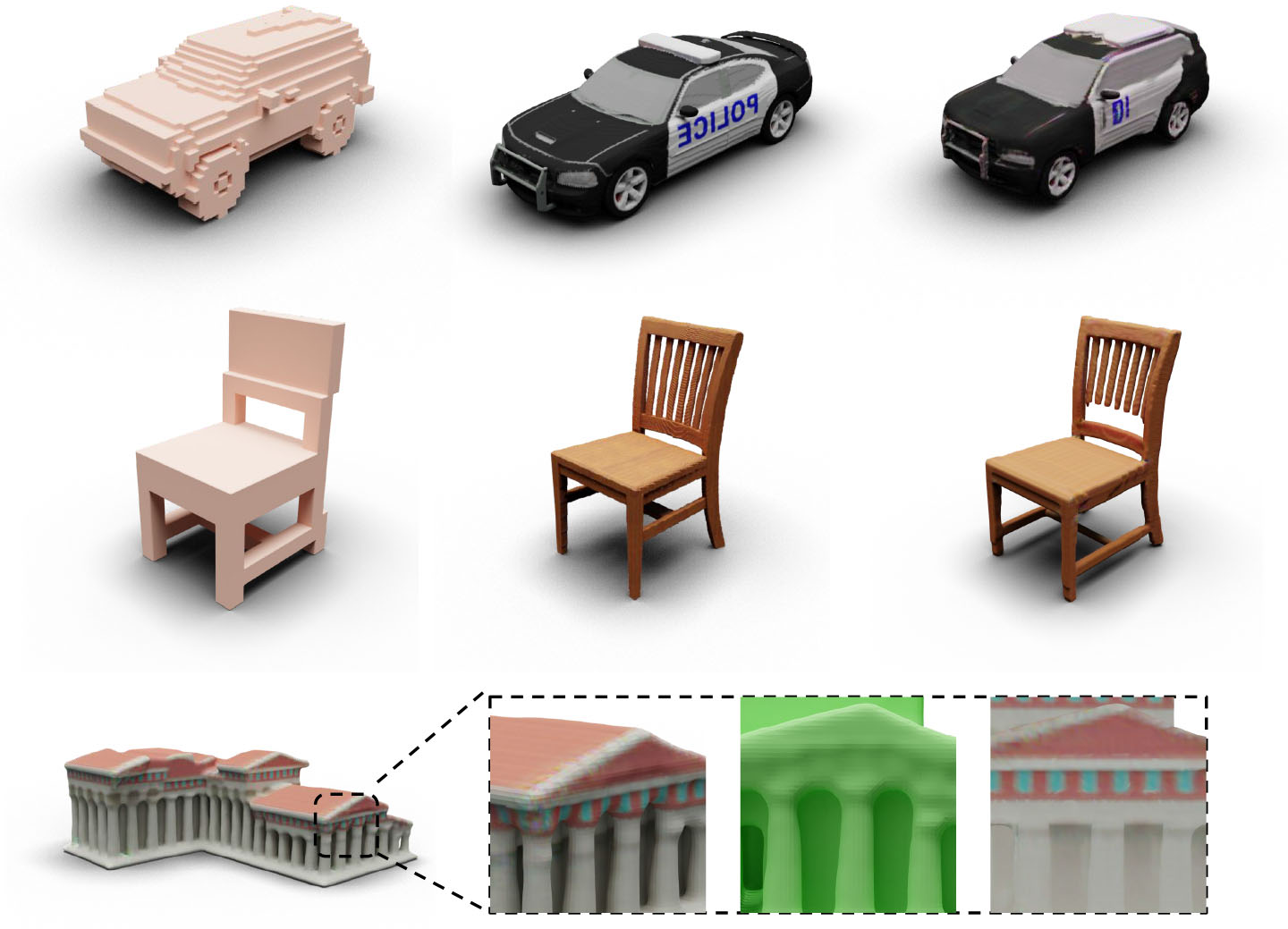}}
  \put(0, 155){\small (a)}
  \put(23, 133){\footnotesize Coarse content}
  \put(107, 133){\footnotesize Style shape}
  \put(180, 133){\footnotesize Our detailization}
  
  \put(0, 95){\small (b)}
  \put(23, 63){\footnotesize Coarse content}
  \put(107, 63){\footnotesize Style shape}
  \put(180, 63){\footnotesize Our detailization}
  
  \put(0, 32){\small (c)}
  \put(20, 3){\footnotesize Our detailization}
  \put(100, 3){\footnotesize Zoom-in}
  \put(139, 3){\footnotesize Geometry only}
  \put(189, 3){\footnotesize Texture only}
\end{picture}
  \vspace{-7mm}
  \caption{Limitations. See Figure~\ref{fig:teaser} for the geometry and texture styles of (c).}
  \vspace{-3mm}
  \label{fig:limitation}
\end{figure}
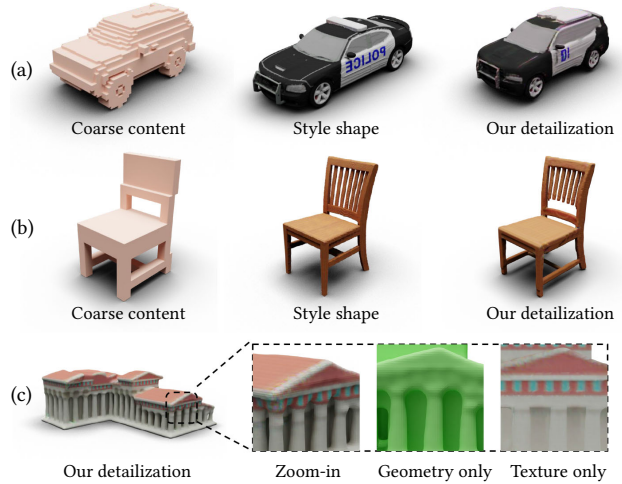

We present \ApproachName, an example-based deep generative network that produces a high-resolution textured 3D shape through geometry detailization
and conditional texture generation. \ApproachName is capable of upsampling a coarse voxel model into a fully textured and geometrically detailed 3D shape, whose results show improved geometric details and clean textures compared to prior works. Extensive experiments demonstrate the capability of \ApproachName to generate novel detailed textured shapes, whether trained on synthetic or real data. The interactive modeling interface powered by \ApproachName allows users to edit the coarse content voxel model and visualize the detailed textured shape in real time.

\emph{Limitations}. 
As our texture generation is designed to only enforce the plausibility of local patches, the generated textures may lack global structure awareness. For instance, the detailized car in Figure \ref{fig:limitation} (a) does not possess the same blue letters as those in the style shape, which can be explained by considering that both blue letters and plain white are plausible textures presented in the style shape.
Similarly, our geometry detailization relies solely on generating plausible local patches, which may lead to global inconsistencies in the output shapes. An example is shown in Figure \ref{fig:limitation} (b), where a portion of the back is carved out in the detailized chair to respect the structure of the content shape, leaving inconsistent edges.
In addition, our method does not have an explicit mechanism to enforce the alignment between the generated textures and the underlying geometry, thus it may lead to geometry-texture misalignment, as in Figure \ref{fig:limitation} (c). However, this phenomenon is surprisingly rare in our experiments, which may be attributed to our texture discriminators ensuring multi-view plausibility of the synthesized textures.

\emph{Future works}.
Aside from addressing the limitations above, we are interested in learning both geometry detailization and texture generation without any 3D supervision, possibly with differentiable volumetric rendering. Introducing local control is another interesting direction, e.g., allowing users to specify arbitrary regions of the coarse content shape and detailize them into designated styles.

\section*{Acknowledgments}

We thank all the anonymous reviewers for their insightful comments and constructive feedback. This work was supported in part by an NSERC Discovery Grant (No.~611370), Adobe gift funds, and
a Google PhD Fellowship received by the second author.

\bibliographystyle{ACM-Reference-Format}
\bibliography{bibliography}

\newpage
\begin{figure*}
\begin{picture}(510, 180)
  \put(0, 0){\includegraphics[width=0.99\linewidth]{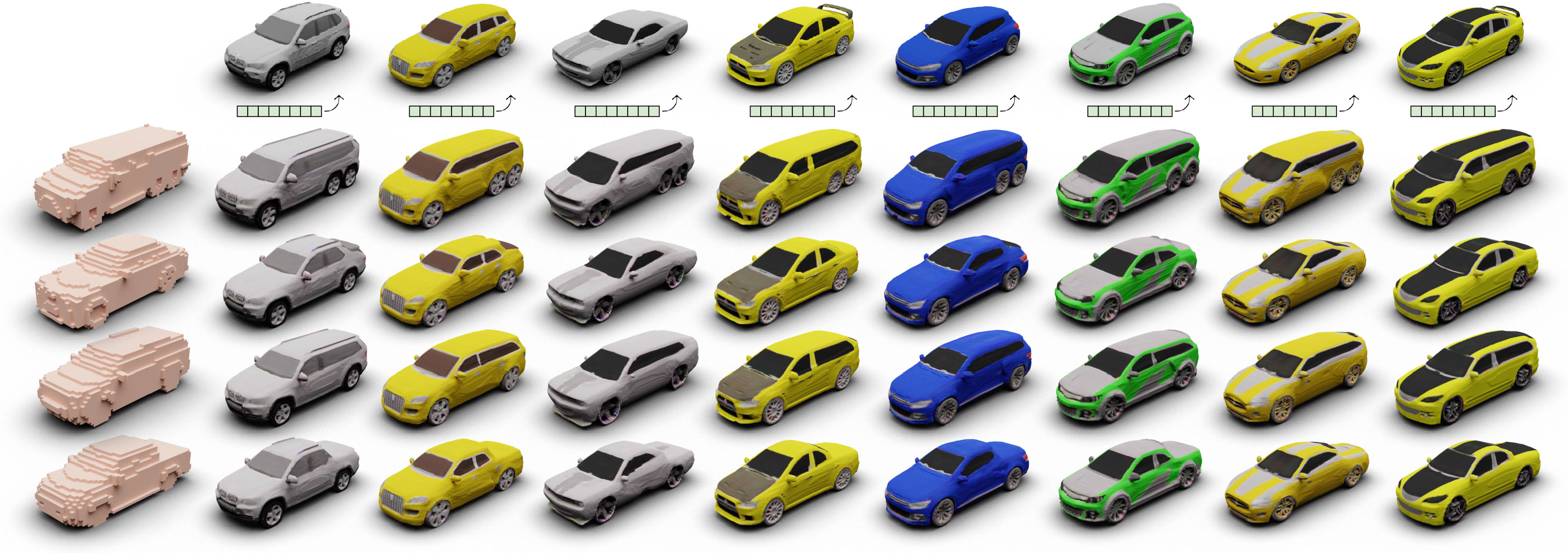}}
  \linethickness{0.005mm}
  \put(15, 170){\line(20,-8){48}}
  \put(15, 150){\small Content}
  \put(39, 168){\small Texture}
  \put(47, 160){\small Code}
  \end{picture}
  \vspace{-6mm}
  \caption{Results of geometry detailization and texture generation on the car category. We show the input coarse content voxels on the left and the detailed style shapes with textures on top. The coarse content voxels are $64^3$ and the generated shapes are $512^3$.}
  \label{fig:car_results}
\end{figure*}
\begin{figure*}
\begin{picture}(510, 167)
  \put(0, 0){\includegraphics[width=0.99\linewidth]{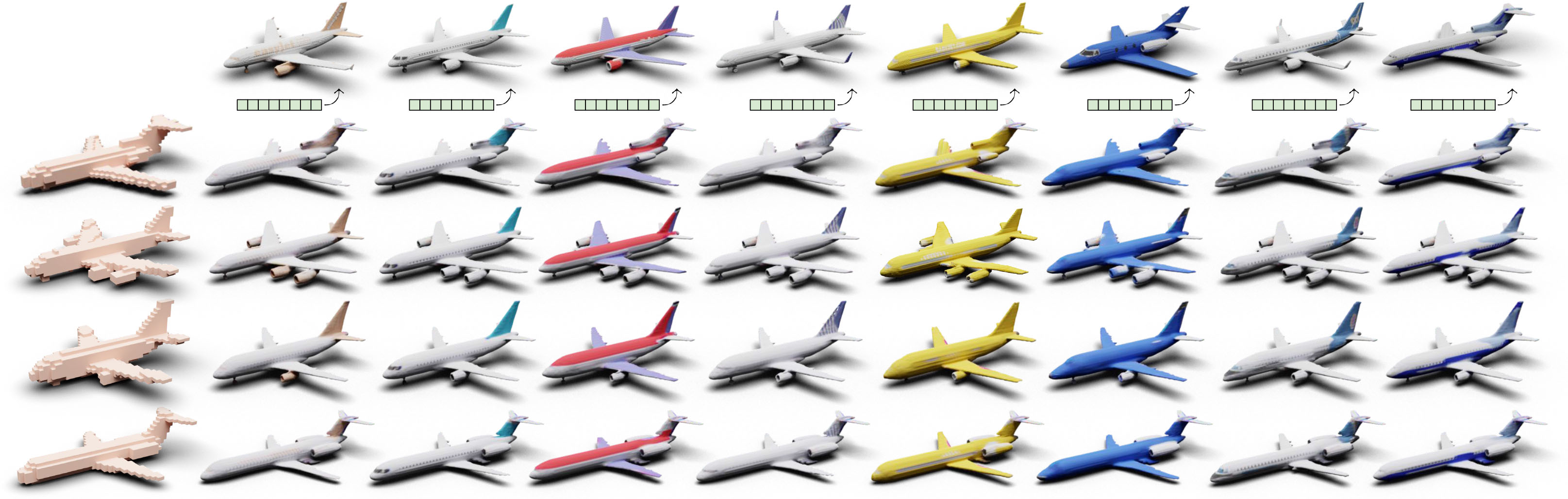}}
  \linethickness{0.005mm}
  \put(10,153){\line(15,-6){48}}
  \put(10, 133){\small Content}
  \put(34, 150){\small Texture}
  \put(42, 143){\small Code}
  \end{picture}
  \vspace{-6mm}
  \caption{Results of geometry detailization and texture generation on the airplane category. We show the input coarse content voxels on the left and the detailed style shapes with textures on top. The coarse content voxels are $64^3$ and the generated shapes are $512^3$.}
  \label{fig:plane_results}
\end{figure*}
\begin{figure*}
\begin{picture}(510, 170)
  \put(0, 0){\includegraphics[width=0.99\linewidth]{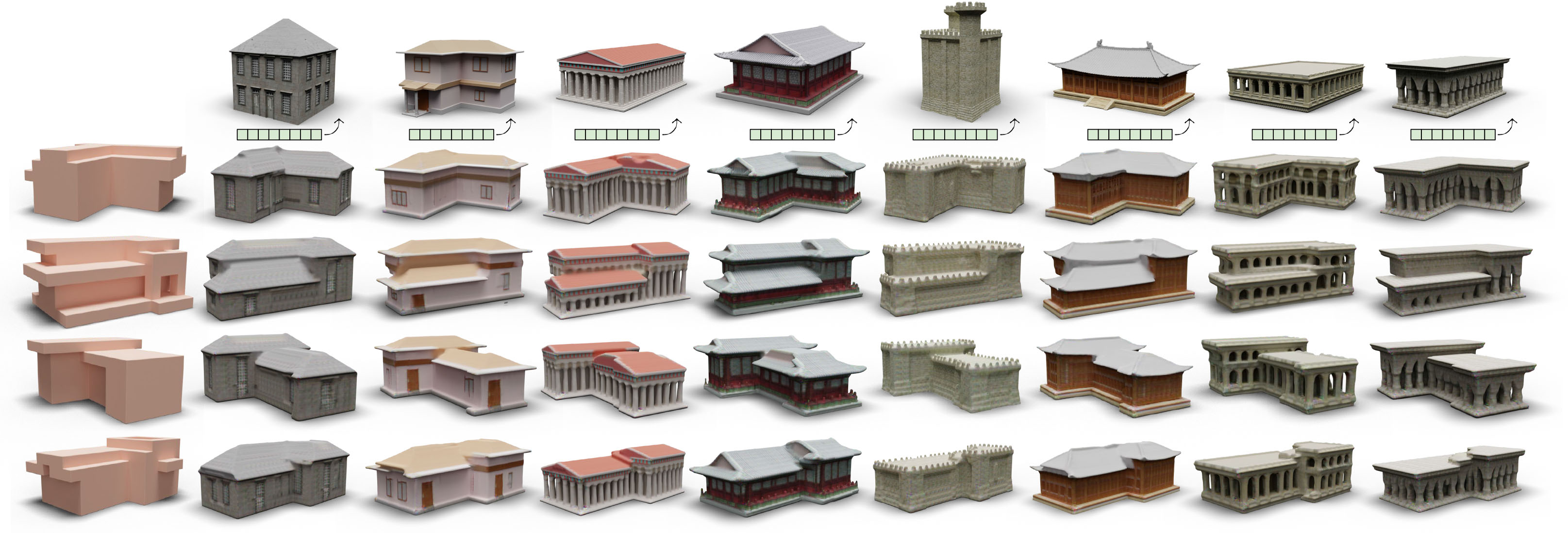}}
  \linethickness{0.005mm}
  \put(8, 160){\line(13,-5){48}}
  \put(8, 140){\small Content}
  \put(32, 160){\small Texture}
  \put(40, 150){\small Code}
\end{picture}
\vspace{-6mm}
  \caption{Results of geometry detailization and texture generation on the building category. We show the input coarse content voxels on the left and the detailed style shapes with textures on top. The coarse content voxels are $32^3$ and the generated shapes are $256^3$.}
  \label{fig:building_results}
\end{figure*}
\begin{figure*}
\begin{picture}(510, 255)
  \put(0, 0){\includegraphics[width=0.99\linewidth]{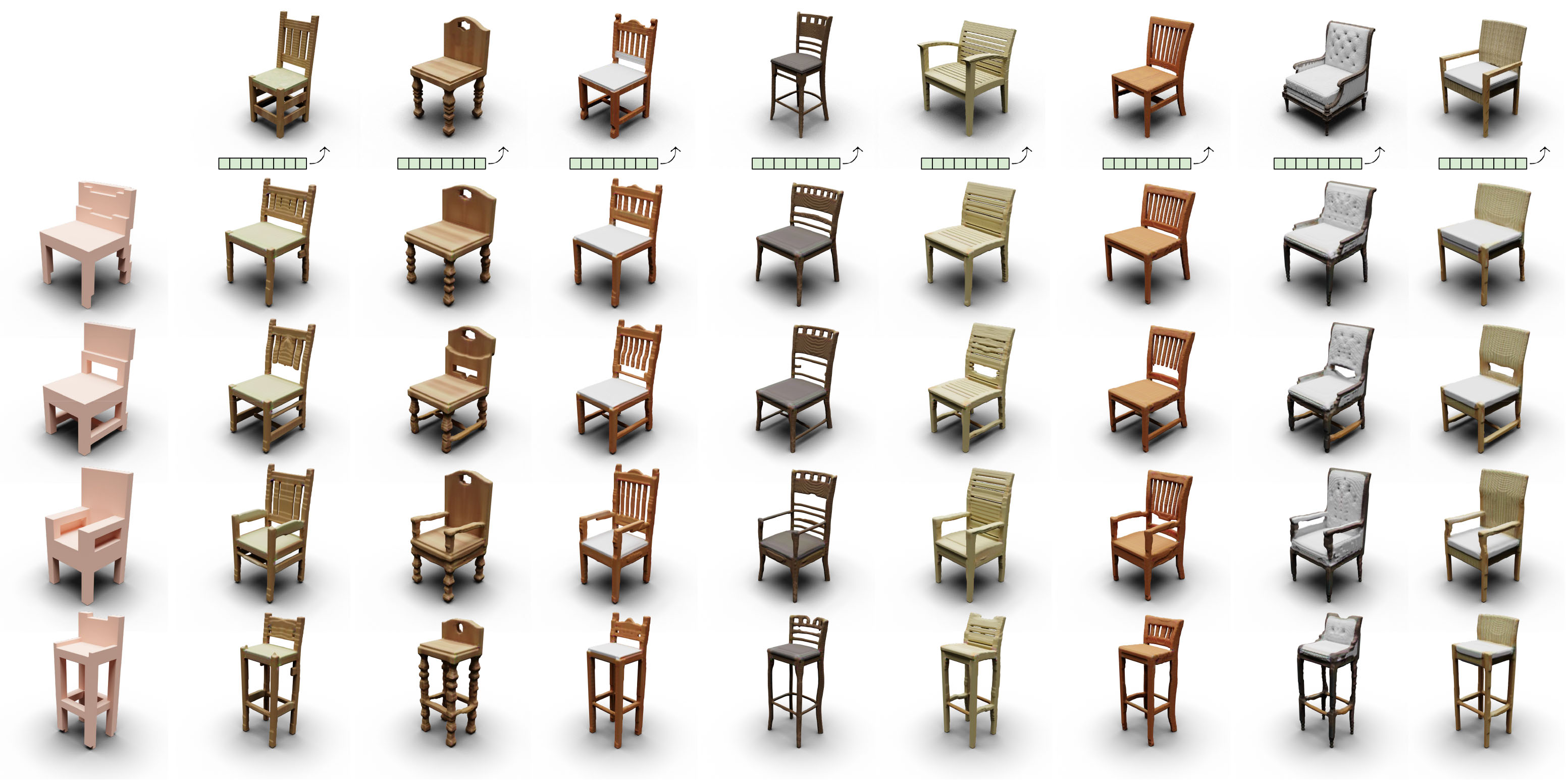}}
  \linethickness{0.005mm}
  \put(5,235){\line(10,-4){48}}
  \put(5, 215){\small Content}
  \put(29, 233){\small Texture}
  \put(37, 225){\small Code}
\end{picture}
  \vspace{-5mm}
  \caption{Results of geometry detailization and texture generation on the chair category.  We show the input coarse content voxels on the left and the detailed style shapes with textures on top. The coarse content voxels are $32^3$ and the generated shapes are $256^3$.}
  \label{fig:chair_results}
\end{figure*}
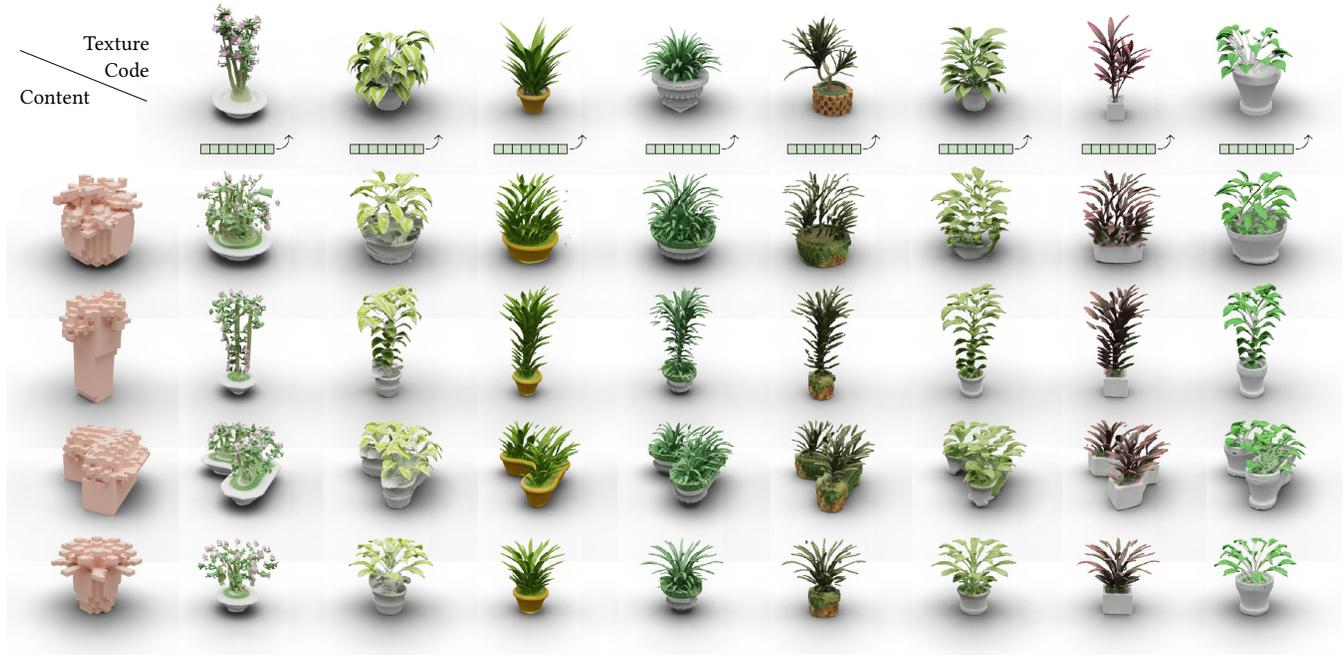
\begin{figure*}
\begin{picture}(510, 255)
  \put(0, 0){\includegraphics[width=0.99\linewidth]{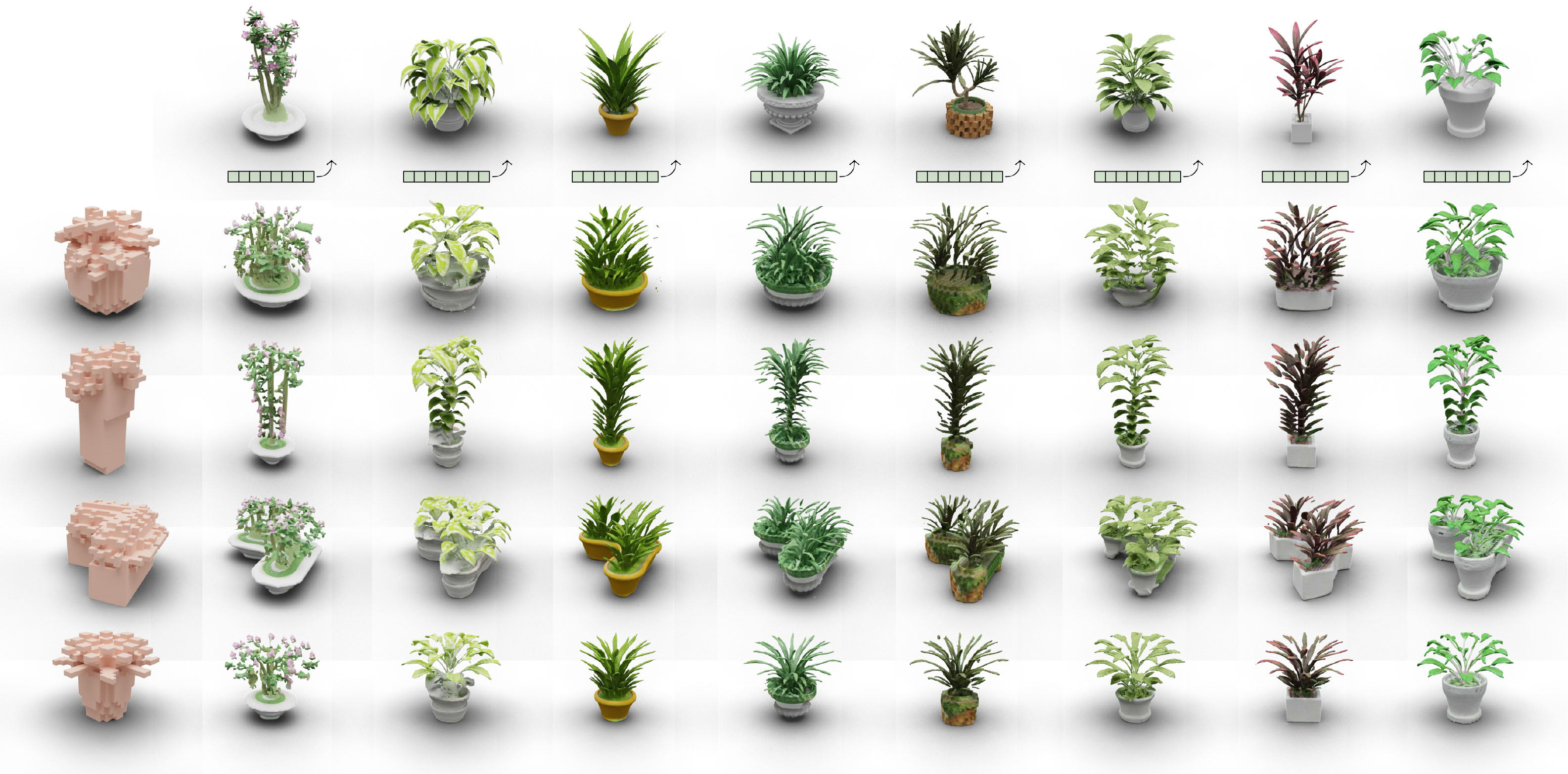}}
  \linethickness{0.005mm}
  \put(5,230){\line(10,-4){48}}
  \put(5, 210){\small Content}
  \put(29, 230){\small Texture}
  \put(37, 220){\small Code}
\end{picture}
  \vspace{-5mm}
  \caption{Results of geometry detailization and texture generation on the plant category.  We show the input coarse content voxels on the left and the detailed style shapes with textures on top. The coarse content voxels are $32^3$ and the generated shapes are $256^3$.}
  \label{fig:plant_results}
\end{figure*}

\clearpage
\newpage
\twocolumn[
\centering
\textbf{\Large{Supplementary Material}} \\
\vspace{2.0em}
]

In this supplementary material, we provide details regarding network architectures, loss functions, implementation settings and data preparation. We also provide additional qualitative results.

\section{Network Architecture}

\begin{figure*}
  \includegraphics[width=0.95\linewidth]{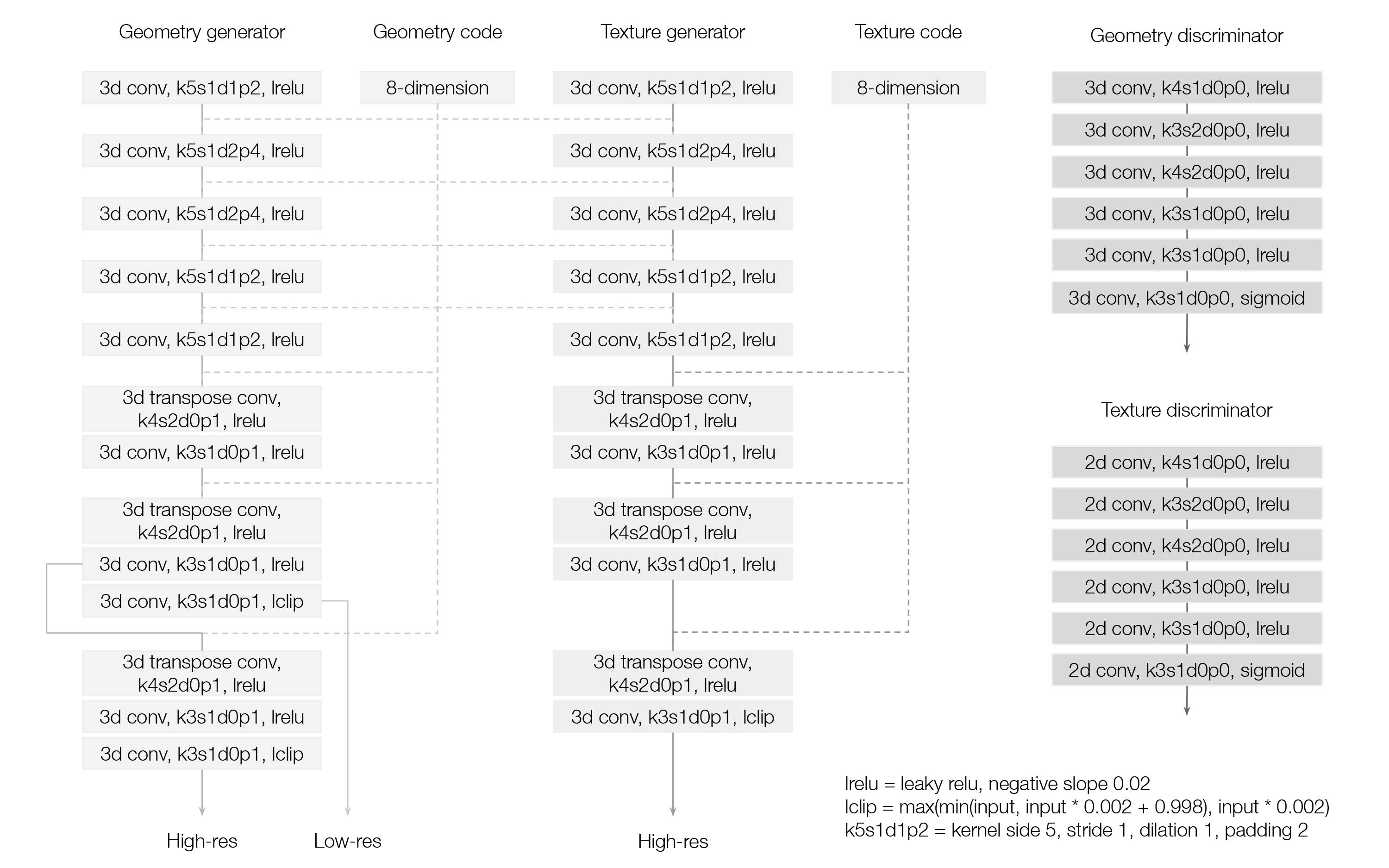}
  \vspace{-3mm}
  \caption{Detailed network architecture of the geometry generator, the texture generator, the geometry discriminator and the texture discriminator.}
  \label{fig:detailed_network_architecture}
\end{figure*}

We provide detailed network architecture in Figure \ref{fig:detailed_network_architecture}.

\emph{Geometry generator}. The geometry generator consists of a backbone network and an upsampling network. The backbone network consists of 5 layers of 3D convolution followed by leaky ReLU. The output of each convolution layer is concatenated with an 8-dimensional trainable module representing the geometry style. The upsampling network consists of 3 upsampling layers. Each upsampling layer is followed by a 3D convolution and doubles the input resolution. The output of the first two upsampling layers is concatenated with the same 8-dimensional trainable module as well. Note that the second upsampling layer is followed by an output layer to output the upsampled shape at intermediate resolution.

\emph{Texture generator}. Similar to the geometry generator, the texture generator consists of a backbone network and an upsampling network. The backbone network consists of 5 layers of 3D convolution followed by leaky ReLU. The output of each convolution layer is concatenated with an 8-dimensional \textit{pre-trained} module representing the geometry style learned during geometry detailization. The upsampling network consists of 3 upsampling layers. Each upsampling layer is followed by a 3D convolution and doubles the input resolution. The output of the first two upsampling layers is concatenated with an 8-dimensional trainable module representing the texture style.

\emph{Geometry discriminator}. The geometry discriminator consists of 5 layers of \textit{3D} convolution followed by leaky ReLU and a layer of 3D convolution followed by Sigmoid. The receptive fields of the geometry discriminator are adjusted according to the category. We use receptive field $36\times36\times36$ for car and plane and $18\times18\times18$ for chair, building and plant.

\emph{Texture discriminator}. The geometry discriminator consists of 5 layers of \textit{2D} convolution followed by leaky ReLU and a layer of 3D convolution followed by Sigmoid. The receptive fields of the texture discriminator are adjusted according to the category. We use receptive field $36\times36$ for car and plane and $18\times18$ for chair, building and plant.

\section{Loss function}

Like DECOR-GAN, We provide a detailed explanation of annotations defined in Equations 1, 2, 3, 4, 5, 6, 7, 8.

\emph{$G_{K}^{geo}$}. The geometry generator that outputs the geometry voxel of resolution $K^{3}$.

\emph{$D_{K}^{geo}$}. The geometry discriminator that determines the patches of the geometry voxel at the resolution $K^{3}$. More specifically, $^{global}{D}_{K}^{geo}$ denotes the global branch of the geometry discriminator at the resolution $K^{3}$ and $^{style}D_{K}^{geo}$ denotes the style branch of the geometry discriminator at the resolution $K^{3}$.

\emph{$G^{tex}$}. The texture generator that outputs the volumetric texture of resolution $K^{3}$.

\emph{$D_{i}^{tex}$}. The texture discriminator that determines the patches of the rendered image at view $i$. More specifically, $^{global}D_{i}^{tex}$ denotes the global branch of the texture discriminator at view $i$ and $^{style}D_{i}^{tex}$ denotes the style branch of the texture discriminator at view $i$.

\emph{$M_{c\cdot K}^{G}$}. The generator mask for content shape. This mask ensures that the empty voxel in the input coarse content shape remains empty in the output upsampled shape and allows the generator to focus on generating voxels in the valid region. It is computed by dilating the coarse input content and upsampling it by $8$ times.

\emph{$M_{s\cdot K}^{G}$}. Similar to $M_{c\cdot K}^{G}$. It is computed by first dilating the downsampled style shape and upsampling it by $8$ times.

\emph{$M_{c\cdot K}^{D}$}. The discriminator mask for upsampled shape. This mask ensures that the occupied voxel in the input coarse content shape should lead to the creation of upsampled voxel in its corresponding area. Empty voxels in the region of interest should be punished by the discriminator. It is computed by upsampling the coarse content shape to match the dimension of the discriminator output. It is defined in 2D space for texture generation.

\emph{$M_{s\cdot K}^{D}$}. Similar to $M_{c\cdot K}^{D}$. It is computed by upsampling the downsampled style shape to match the dimension of the discriminator output. It is defined in 2D space for texture generation.

\emph{$c_{s\cdot K}^{geo}$}. The output of the geometry generator. This is the upsampled shape to be discriminated by the geometry discriminator. It is also used for rendering texture images.

\emph{$c_{s\cdot K}^{tex}$}. The output of the texture generator. This is the generated texture voxel to be used for rendering texture images.

\emph{$R_{i}(c_{s\cdot K}^{geo}, c_{s\cdot K}^{tex})$}. The rendering function for view $i$. The rendering function takes upsampled geometry voxel and generated texture voxel as inputs. The 2D masks and depth maps of different views are computed and the rendered images of each view are gathered from the texture voxel. Take the front view for an example, it first calculates the mask (maximum value) and depth (index of the maximum value) by computing the argmax along the first axis of the upsampled geometry voxel, the depth value is then used to gather the color value from the generated texture voxel along the first axis. The mask is then multiplied to mask out the non-region of interest.

\emph{Detailed loss functions}. We follow DECOR-GAN to prevent model collapse by splitting the discriminator into $N+1$ branches at the output layer, where $N$ is the number of detailed shapes and an additional $1$ branch for the global discriminator accounting for all styles. Note that since both style branches and global branch use the exact same equation to compute loss, we omit the details in the main paper for simplicity and clarity. Here we describe the detailed version of loss functions (1), (2), (5) and (6) of the main paper.

For geometry generation, the discriminator loss consists of the global branch's loss ${}^{global}D_{K}^{geo}$ and the style branch's loss ${}^{style}D_{K}^{geo}$ at the resolution of $(\nicefrac{K}{2})^{3}$ and $K^{3}$, we define losses at the resolution of $K^{3}$ here, losses at the resolution of $(\nicefrac{K}{2})^{3}$ can be easily derived by changing the subscript:
\begin{equation}
    \mathcal{L}_{D}^{geo} = \mathcal{L}_{D\cdot K}^{global} + \mathcal{L}_{D\cdot K}^{style} + \mathcal{L}_{D\cdot \nicefrac{K}{2}}^{global} + \mathcal{L}_{D\cdot \nicefrac{K}{2}}^{style}
\end{equation}
where
\begin{align}
    \mathcal{L}_{D\cdot K}^{global} &= \underset{s\sim \mathcal{S}}{\mathbb{E}}\frac{||({}^{global}D_{K}^{geo}(s^{geo})-1)\circ M_{s\cdot K}^{D}||_{2}^{2}}{||M_{s\cdot K}^{D}||_{1}} \nonumber \\ &+ \mathop{\underset{s\sim \mathcal{S}}{\mathbb{E}}}_{c\sim \mathcal{C}}\frac{||{}^{global}D_{K}^{geo}(c_{s\cdot K}^{geo})\circ M_{c\cdot K}^{D}||_{2}^{2}}{||M_{c\cdot K}^{D}||_{1}}\;,\nonumber
\end{align}
\begin{equation}
\label{eqn:geo_dis_loss}
    c_{s\cdot K}^{geo} = G_{K}^{geo}(c,\, z_{s}^{geo})\circ M_{c\cdot K}^{G}
\end{equation}

\begin{align}
    \mathcal{L}_{D\cdot K}^{style} &= \underset{s\sim \mathcal{S}}{\mathbb{E}}\frac{||({}^{style}D_{K}^{geo}(s^{geo})-1)\circ M_{s\cdot K}^{D}||_{2}^{2}}{||M_{s\cdot K}^{D}||_{1}}  \nonumber \\ &+ \mathop{\underset{s\sim \mathcal{S}}{\mathbb{E}}}_{c\sim \mathcal{C}}\frac{||{}^{style}D_{K}^{geo}(c_{s\cdot K}^{geo})\circ M_{c\cdot K}^{D}||_{2}^{2}}{||M_{c\cdot K}^{D}||_{1}}
\end{align}
and $\circ$ denotes element-wise multiplication, and $c_{s\cdot K}^{geo}$ is the upsampled shape of resolution $K^3$ from input coarse shape $c$ with the style of $s$.
The generator loss is defined as:
\begin{equation}
    \mathcal{L}_{G}^{geo} = (\mathcal{L}_{G\cdot K}^{global} + \alpha \cdot \mathcal{L}_{G\cdot K}^{style}) + \gamma\cdot(\mathcal{L}_{G\cdot \nicefrac{K}{2}}^{global} + \alpha \cdot \mathcal{L}_{G\cdot \nicefrac{K}{2}}^{style})
\end{equation}
where
\begin{equation}
    \mathcal{L}_{G\cdot K}^{global} = \mathop{\underset{s\sim \mathcal{S}}{\mathbb{E}}}_{c\sim \mathcal{C}}\frac{||({}^{global}D_{K}^{geo}(c_{s\cdot K}^{geo})-1)\circ M_{c\cdot K}^{D}||_{2}^{2}}{||M_{c\cdot K}^{D}||_{1}}
\end{equation}

\begin{equation}
    \mathcal{L}_{G\cdot K}^{style} = \mathop{\underset{s\sim \mathcal{S}}{\mathbb{E}}}_{c\sim \mathcal{C}}\frac{||({}^{style}D_{K}^{geo}(c_{s\cdot K}^{geo})-1)\circ M_{c\cdot K}^{D}||_{2}^{2}}{||M_{c\cdot K}^{D}||_{1}}
\end{equation}
With the reconstruction loss described in the Section 3.1, the overall generator loss for geometry detailization is:
\begin{align}
    \mathcal{L}_{geo} = \mathcal{L}_{G}^{geo} + \beta \cdot \mathcal{L}_{K}^{recon} + \beta \cdot \mathcal{L}_{\nicefrac{K}{2}}^{recon}
\end{align}

For texture generation, the discriminator loss consists of the global branch's loss ${}^{global}D_{i}^{tex}$ and the style branch's loss ${}^{style}D_{i}^{tex}$.
\begin{equation}
    \mathcal{L}_{D}^{tex} = \sum_{i}(\mathcal{L}_{D\cdot i}^{global} + \mathcal{L}_{D\cdot i}^{style})
\end{equation}
where
\begin{align}
    \mathcal{L}_{D\cdot i}^{global} &= \underset{s\sim \mathcal{S}}{\mathbb{E}}\frac{||({}^{global}D_{i}^{tex}(R_{i}(s^{geo},\, s^{tex}))-1)\circ M_{s\cdot i}^{D}||_{2}^{2}}{||M_{s\cdot i}^{D}||_{1}} \nonumber \\ & + \mathop{\underset{s\sim \mathcal{S}}{\mathbb{E}}}_{c\sim \mathcal{C}}\frac{||{}^{global}D_{i}^{tex}(R_{i}(c_{s\cdot K}^{geo},\, c_{s\cdot K}^{tex})\circ M_{c\cdot i}^{D}||_{2}^{2}}{||M_{c\cdot i}^{D}||_{1}}\;, \nonumber
\end{align}
\begin{equation}
    c_{s\cdot K}^{tex} = G^{tex}(c,\, z_{s}^{tex})
\end{equation}
\begin{align}
    \mathcal{L}_{D\cdot i}^{style} &= \underset{s\sim \mathcal{S}}{\mathbb{E}}\frac{||({}^{style}D_{i}^{tex}(R_{i}(s^{geo},\, s^{tex}))-1)\circ M_{s\cdot i}^{D}||_{2}^{2}}{||M_{s\cdot i}^{D}||_{1}} \nonumber \\ & + \mathop{\underset{s\sim \mathcal{S}}{\mathbb{E}}}_{c\sim \mathcal{C}}\frac{||{}^{style}D_{i}^{tex}(R_{i}(c_{s\cdot K}^{geo},\, c_{s\cdot K}^{tex})\circ M_{c\cdot i}^{D}||_{2}^{2}}{||M_{c\cdot i}^{D}||_{1}}
\end{align}
where $c_{s\cdot K}^{tex}$ is the synthesized color grid, and $R_{i}(\cdot, \cdot)$ is the rendering function at view $i$ given the geometry voxels and the color grid. The generator loss is defined as:
\begin{equation}
    \mathcal{L}_{G\cdot i}^{tex} = \mathcal{L}_{G\cdot i}^{global} + \gamma_{1} \cdot \mathcal{L}_{G\cdot i}^{style}
\end{equation}
where
\begin{equation}
    \mathcal{L}_{G\cdot i}^{global} = \mathop{\underset{s\sim \mathcal{S}}{\mathbb{E}}}_{c\sim \mathcal{C}}\frac{||({}^{global}D_{i}^{tex}(R_{i}(c_{s\cdot K}^{geo},\, c_{s\cdot K}^{tex}))-1)\circ M_{c\cdot i}^{D}||_{2}^{2}}{||M_{c\cdot i}^{D}||_{1}}
\end{equation}
\begin{equation}
    \mathcal{L}_{G\cdot i}^{style} = \mathop{\underset{s\sim \mathcal{S}}{\mathbb{E}}}_{c\sim \mathcal{C}}\frac{||({}^{style}D_{i}^{tex}(R_{i}(c_{s\cdot K}^{geo},\, c_{s\cdot K}^{tex}))-1)\circ M_{c\cdot i}^{D}||_{2}^{2}}{||M_{c\cdot i}^{D}||_{1}}
\end{equation}
With the reconstruction loss described in the section 3.2, the overall generator loss for texture generation is:
\begin{equation}
    \mathcal{L}_{tex} = \sum_{i}(\mathcal{L}_{G\cdot i}^{tex} + \gamma_{2} \cdot \mathcal{L}_{i}^{recon})
\end{equation}

Empirically we found that using the same value of $\gamma_{1}$ and $\gamma_{2}$ as geometry detailization is enough to obtain good results.

\section{Implementation Details}

We provide data preparation details and hyper-parameters used in all experiments.

\emph{Volumetric textures generation}. In order to generate volumetric texture voxels for detailed style shapes, we first render images of resolution $K\times K\times 4$ from different views of the detailed style mesh, we then take the geometry voxel, approximate the normal direction of each occupied voxel grid, and record surface voxels. We determine the color of each surface voxel by finding the angle between the surface normal and horizontal or vertical plane and retrieving the pixel value from the corresponding view, e.g. if the angle between the surface normal of a surface voxel and the horizontal plane is less than $\nicefrac{\pi}{4}$, the corresponding pixel value of the rendered image from right view is painted into that voxel.

\emph{Hyper-parameters}. We set $\alpha=1.0$ for chair, and $\alpha=0.5$ for car, airplane and building. For training texture of the side view, we set $\beta=5.0$ for car and building, $\beta=10.0$ for airplane, and $\beta=1.0$ for chair. We set $\beta=1.0$ for the rest of the views. We set $\gamma=0.5$ in Equation (4) of this supplementary material. We set the batch size to $1$ and the learning rate to $0.0001$ for all experiments. We train individual models for different categories. We train both the geometry detailization and the texture generation for 20 epochs on a single Nvidia GeForce RTX 3090 Ti. Depending on the category, training each model on the geometry detailization and the texture generation takes 12-24 hours and 18-36 hours, respectively.

\emph{Cropping}. To handle the large memory footprint and speed up training, we crop each shape and discard the unoccupied voxel according to its dilated bounding box. For each detailed style shape, we crop the geometry voxel and the texture voxel.

\section{Evaluation metrics}

\begin{table*}[t]
    \begin{center}
    \caption{Quantitative comparison of our multi-resolution geometry upsampling with two baselines on car category.}
    \label{tab:geometry_ablation_car}
    \vspace{-3mm}
    
    \begin{adjustbox}{width=0.99\textwidth}
    \begin{tabular}{rccccccc}
    \toprule
        & Strict-IOU $\uparrow$ & Loose-IOU $\uparrow$ & LP-IOU $\uparrow$ & LP-F-score $\uparrow$ & Div-IoU $\uparrow$ & Div-F-score $\uparrow$ & Cls-score $\downarrow$ \\
    \midrule
       DECOR-GAN & 0.907 & 0.948 & 0.758 & 0.996 & \textbf{0.950} & 0.815 & 0.505 \\
       DECOR-GAN-up & 0.886 & 0.942 & 0.802 & 0.995 & 0.931 & 0.928 & 0.503 \\
       Ours (multi-res) & \textbf{0.908} & \textbf{0.968} & \textbf{0.803} & \textbf{0.997 } & 0.947 & \textbf{0.944} & \textbf{0.502} \\
    \bottomrule
    \end{tabular}
    \end{adjustbox}
    \end{center}
    
\end{table*}

\begin{table*}[t]
    \begin{center}
    \caption{Quantitative comparison of ours multi-resolution geometry upsampling with two baselines on plane category.}
    \label{tab:geometry_ablation_plane}
    \vspace{-3mm}
    
    \begin{adjustbox}{width=0.99\textwidth}
    \begin{tabular}{rccccccc}
    \toprule
        & Strict-IOU $\uparrow$ & Loose-IOU $\uparrow$ & LP-IOU $\uparrow$ & LP-F-score $\uparrow$ & Div-IoU $\uparrow$ & Div-F-score $\uparrow$ & Cls-score $\downarrow$ \\
    \midrule
       DECOR-GAN & 0.813 & 0.919 & 0.462 & 0.995 & 0.484 & 0.347 & 0.513 \\
       DECOR-GAN-up & 0.790 & 0.905 & 0.561 & 0.994 & \textbf{0.665} & \textbf{0.400} & 0.515 \\
       Ours (multi-res) & \textbf{0.839} & \textbf{0.935} & \textbf{0.572} & \textbf{0.996} & 0.606 & 0.366 & \textbf{0.500} \\
    \bottomrule
    \end{tabular}
    \end{adjustbox}
    \end{center}
    
\end{table*}

\begin{table*}[t]
    \begin{center}
    \caption{Quantitative comparison of ours multi-resolution geometry upsampling with two baselines on building category.}
    \label{tab:geometry_ablation_building}
    \vspace{-3mm}
    
    \begin{adjustbox}{width=0.99\textwidth}
    \begin{tabular}{rccccccc}
    \toprule
        & Strict-IOU $\uparrow$ & Loose-IOU $\uparrow$ & LP-IOU $\uparrow$ & LP-F-score $\uparrow$ & Div-IoU $\uparrow$ & Div-F-score $\uparrow$ & Cls-score $\downarrow$ \\
    \midrule
       DECOR-GAN & \textbf{0.823} & 0.904 & 0.739 & 0.964 & 1.000 & 0.944 & 0.520 \\
       DECOR-GAN-up & 0.765 & 0.903 & 0.720 & 0.959 & 1.000 & 0.987 & \textbf{0.513} \\
       Ours (multi-res) & 0.811 & \textbf{0.907} & \textbf{0.740} & \textbf{0.968} & 1.000 & \textbf{1.000} & 0.545 \\
    \bottomrule
    \end{tabular}
    \end{adjustbox}
    \end{center}
    
\end{table*}

We adopt the metrics from DECOR-GAN to quantitatively evaluate the quality of the generated geometry and texture.

\emph{Strict-IoU and Loose-IoU.} We use Strict-IoU to evaluate how much the generated shape respects the input coarse content shape since the downsampled version of the output should ideally be identical to the input. Since we use dilated generator mask so that the generator can generate shapes within dilated regions, we also use Loose-IOU between downsampled output voxels and the input voxels by discarding the voxels in the dilated portion of the input.

\emph{LP-IoU and LP-F-score.} We use the Local Plausibility (LP) with two measuring distances (IoU or F-score) to evaluate the percentage of local patches in the output shape that are ``similar'' to at least one local patch in the detailed style shapes. For LP-IoU, two patches are considered to be ``similar'' if the IoU is above 0.95. For LP-F-score, two patches are considered to be ``similar'' if the F-score is above 0.95. We use the same implementation as DECOR-GAN to compute LP-IoU and LP-F-score.

\emph{Div-IoU and Div-F-score.} We use Div-IoU and Div-F-score to evaluate the diversity of the outputs with respect to the styles. Similar to DECOR-GAN, we denote $N_{ijk}$ as the number of local patches from input voxel $i$, upsampled with style shape $j$, that are "similar" to at least one patch from the detailed style shape $k$. Since the generator might "copy" local patches from different style shapes other than the designated style, which introduces the style bias, we denote $N_{ik}$ to be the mean of $N_{ijk}$ over all possible styles to remove the style bias. We use IoU and F-score as the distance metrics for patches to compute Div-IoU and Div-F-score.

\emph{Cls-score.} We evaluate whether the generated shapes are distinguishable or not from the real shapes by training a ResNet using high-resolution voxels as real data and generated shapes as fake data. We render each voxel sample into $24$ $512^2$ images from random views for the car and plane categories and $24$ $256^2$ images for the chair and building categories. During training, the images are randomly cropped into $64^2$ patches. The mean classification accuracy is denoted as the Cls-score.

\emph{FID-all and FID-style.} Ideally, the generated textures should be globally plausible and locally similar to the ground truth texture providing the style. To quantitatively measure the quality of the textures, we render images of the generated shapes and the style shapes from different views and evaluate with Fréchet Inception Distance (FID). We use FID-all to measure the similarity between the generated textures and all ground truth textures used in training, and FID-style to measure the similarity between the generated textures and the ground truth texture that provides the style of the generated textures. In other words, FID-all will be small as long as the generated texture is similar to any of the style textures in the training set, while in order for FID-style to be small, the generated texture needs to be similar to the texture that provided the style. Thus there is a noticeable scale difference between the two metrics.

\emph{LPIPS-style.} Similar to FID-style, we use LPIPS-style to evaluate the local patch similarity between generated textures and the ground truth textures of each style. \\

\noindent{\textbf{Evaluation details.}} We use the same setting as DECOR-GAN for evaluating the quality of the generated geometry. For LP and Div, we use 20 content shapes from the test set and 16 styles for each content shape, 320 shapes in total. For Cls-score, we use all training shapes as real samples and 100 content shapes from the test set as fake samples with 16 styles for each content shape, 1600 shapes in total. For FID-all, FID-style and LPIPS-style, we use 20 content shapes from the test set and 16 styles for each content shape, 320 shapes in total. For each shape, we uniformly render 24 views by rotating the camera around the object with fixed poses.

\section{Application}

We create a GUI application where users can edit a given coarse content voxel template, including adding or removing voxel cell, choose a style shape and visualize the detailed textured shape in real time. The whole pipeline works as follow:
\begin{enumerate}
    \item User edits the coarse content voxel.
    \item Once user finishes editing, the edited coarse content voxel is passed to the pre-trained model for geometry detailization and texture generation.
    \item Marching Cubes is performed on the upsampled geometry to extract the surface and texture colors are gathered and assigned to each vertex.
    \item The detailed textured shape is visualized on canvas.
\end{enumerate}

We provide instructions on how to use the modeling interface:
\begin{itemize}
    \item Left-click on the computer mouse on the coarse voxel to add voxel(s).
    \item Left-click on the empty space to rotate the camera view.
    \item Right-click on the computer mouse on the coarse voxel to remove voxel(s).
    \item Right-click on the empty space to move the coarse voxel.
    \item Middle-click and scroll the scroll wheel on the computer mouse to zoom in/out.
    \item "Q", "W", "E", "R" for brush sizes "1", "3", "5", "7".
    \item Num pad 1-8 for choosing different styles.
    \item Space for switching between editing and viewing mode.
\end{itemize}

\newpage
\begin{figure*}
  \includegraphics[width=0.97\linewidth]{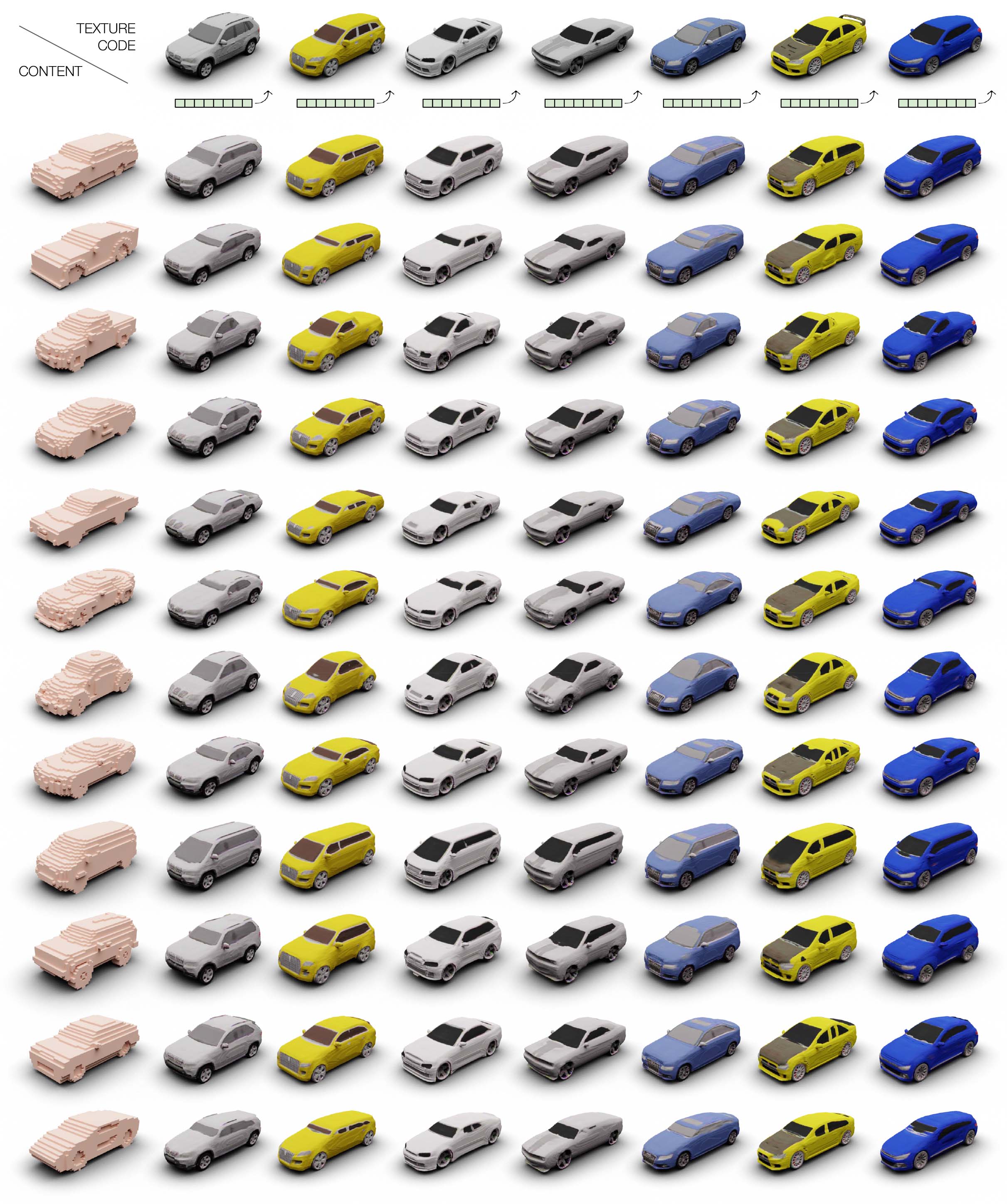}
  \vspace{-4mm}
  \caption{Results of geometry detailization and texture generation on car category. Style shapes are shown in the first row and the input coarse voxels are shown in the first column. The input resolution is $64^{3}$ and the output resolution is $512^{3}$.}
  \label{fig:supp_car_results1}
\end{figure*}

\newpage
\begin{figure*}
  \includegraphics[width=0.95\linewidth]{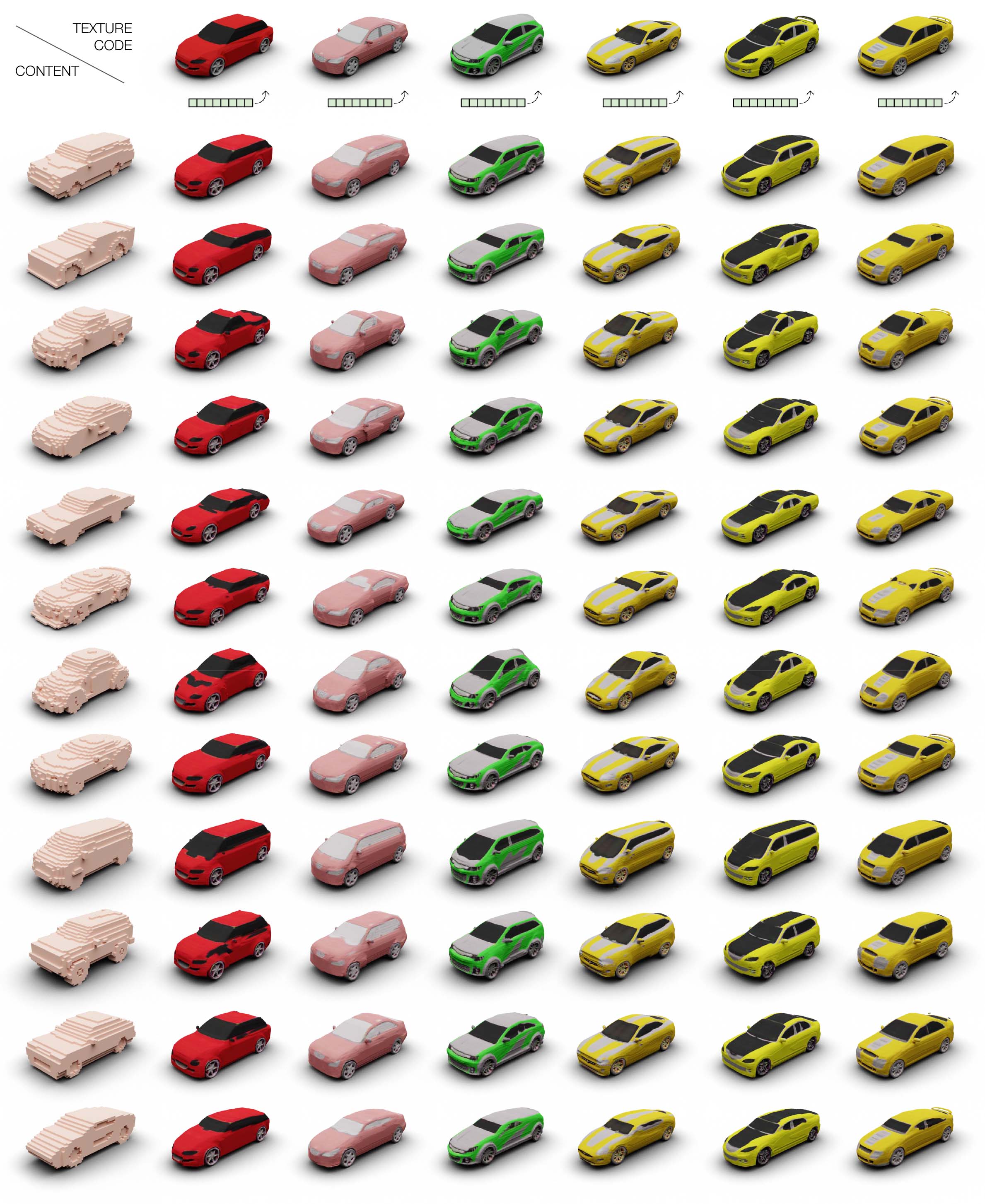}
  \vspace{-4mm}
  \caption{Results of geometry detailization and texture generation on car category. Style shapes are shown in the first row and the input coarse voxels are shown in the first column. The input resolution is $64^{3}$ and the output resolution is $512^{3}$.}
  \label{fig:supp_car_results2}
\end{figure*}

\newpage
\begin{figure*}
\begin{picture}(510,180)
  \put(0, 0){\includegraphics[width=0.99\linewidth]{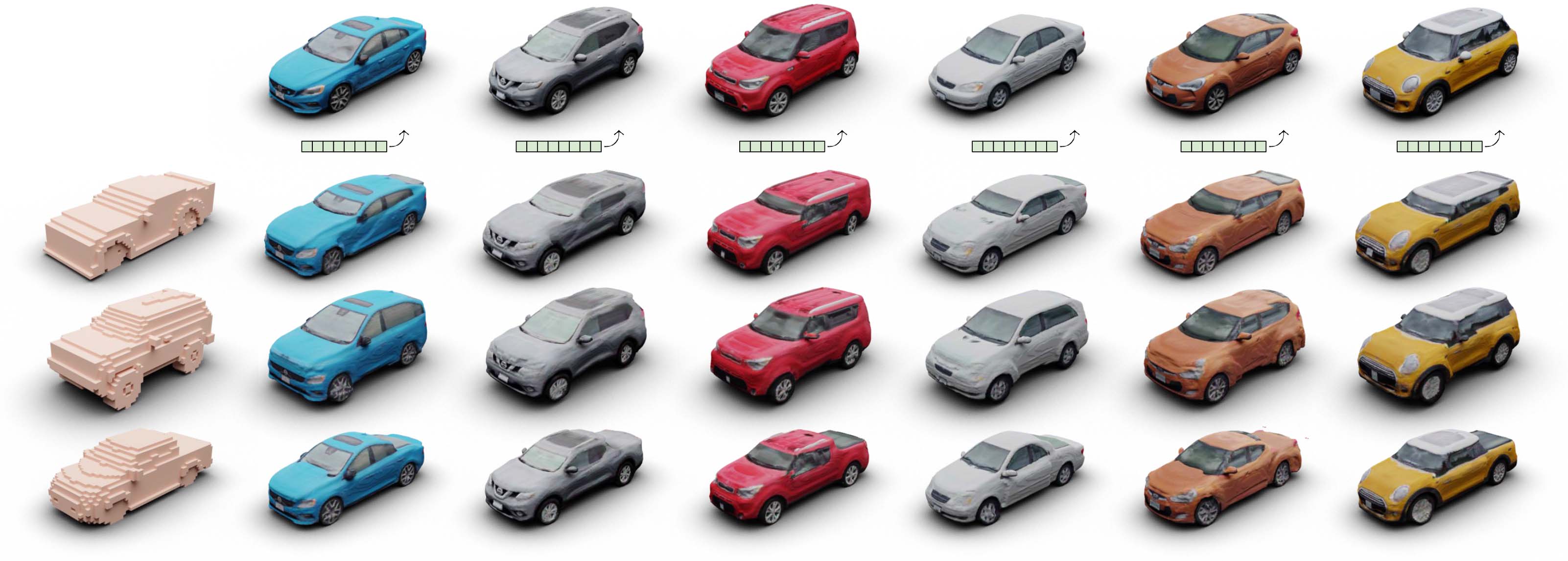}}
  \linethickness{0.005mm}
  \put(15, 165){\line(30, -8){52}}
  \put(15, 148){\small Content}
  \put(45, 168){\small Texture}
  \put(53, 160){\small Code}
\end{picture}
  \vspace{-5mm}
  \caption{Results of geometry detailization and texture generation conditioned on cars that are reconstructed from real-world images. We show the input coarse content voxels on the left and the detailed style shapes with textures on top. The coarse content voxels are $64^3$ and the generated shapes are $512^3$.}
  \label{fig:real_car_results}
\end{figure*}
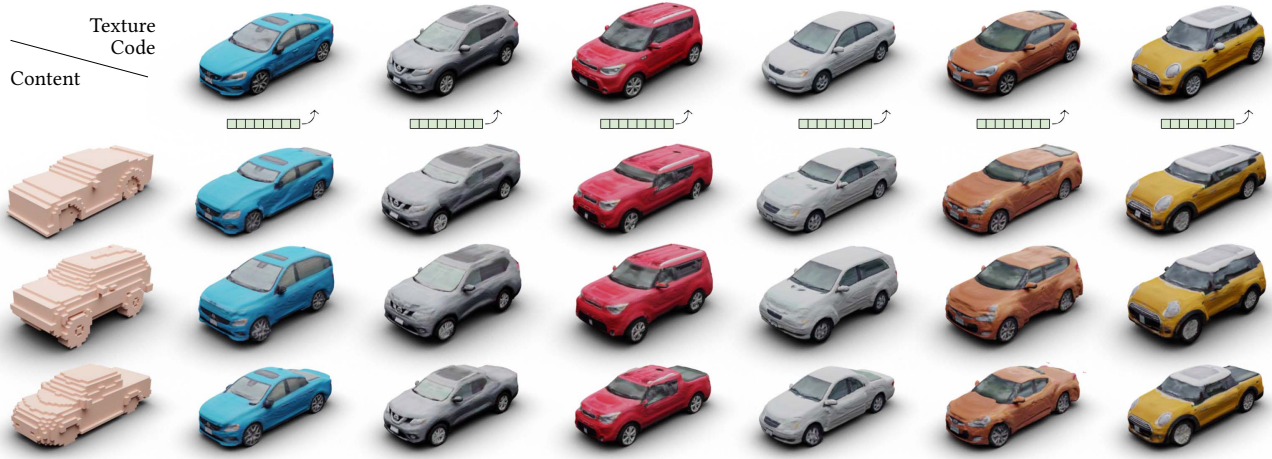

\newpage
\begin{figure*}
  \includegraphics[width=0.92\linewidth]{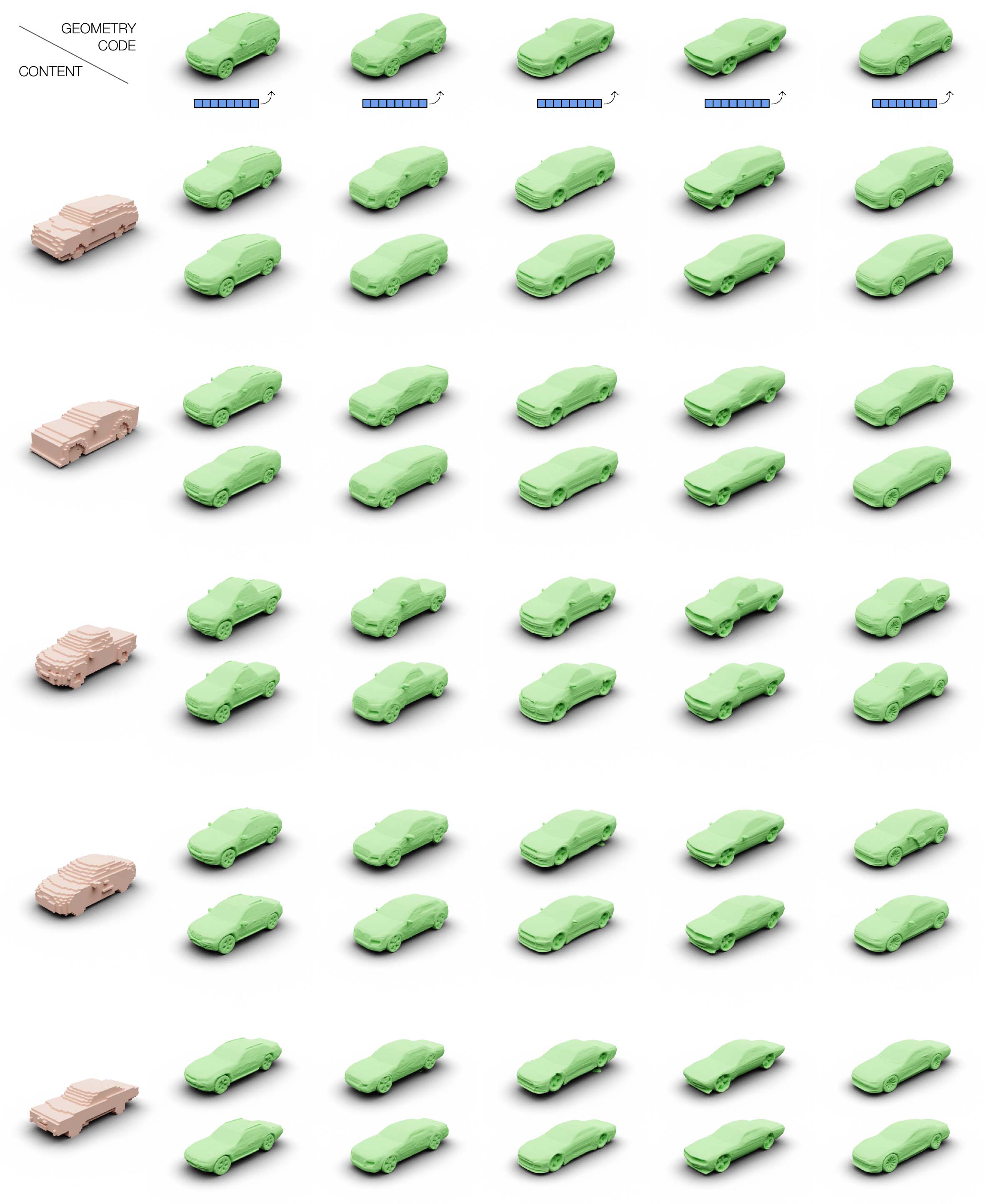}
  \vspace{-2mm}
  \caption{Results of geometry detailization on car category. Style geometries are shown in the first row and the input coarse voxels are shown in the first column. Comparisons of DECOR-GAN-up and our multi-resolution are shown in the first and second row of each coarse voxel, respectively. The input resolution is $64^{3}$ and the output resolution is $512^{3}$. Please zoom in to observe the details.}
  \label{fig:supp_car_results3}
\end{figure*}

\newpage

\begin{figure*}
  \includegraphics[width=0.99\linewidth]{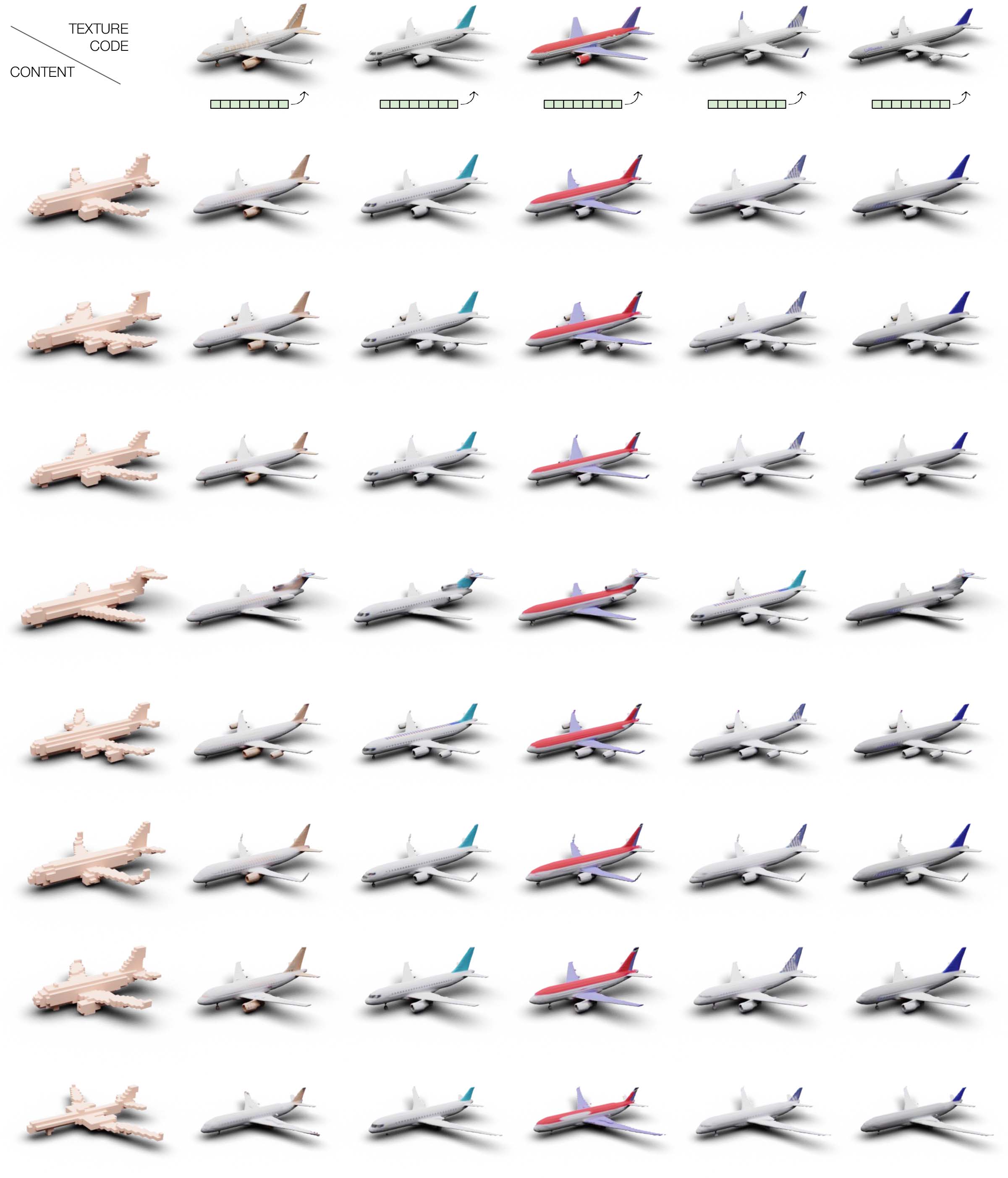}
  \vspace{-4mm}
  \caption{Results of geometry detailization and texture generation on airplane category. Detailed shapes are shown in the first row and the input coarse voxels are shown in the first column. The input resolution is $64^{3}$ and the output resolution is $512^{3}$.}
  \label{fig:supp_plane_results1}
\end{figure*}

\newpage
\begin{figure*}
  \includegraphics[width=0.99\linewidth]{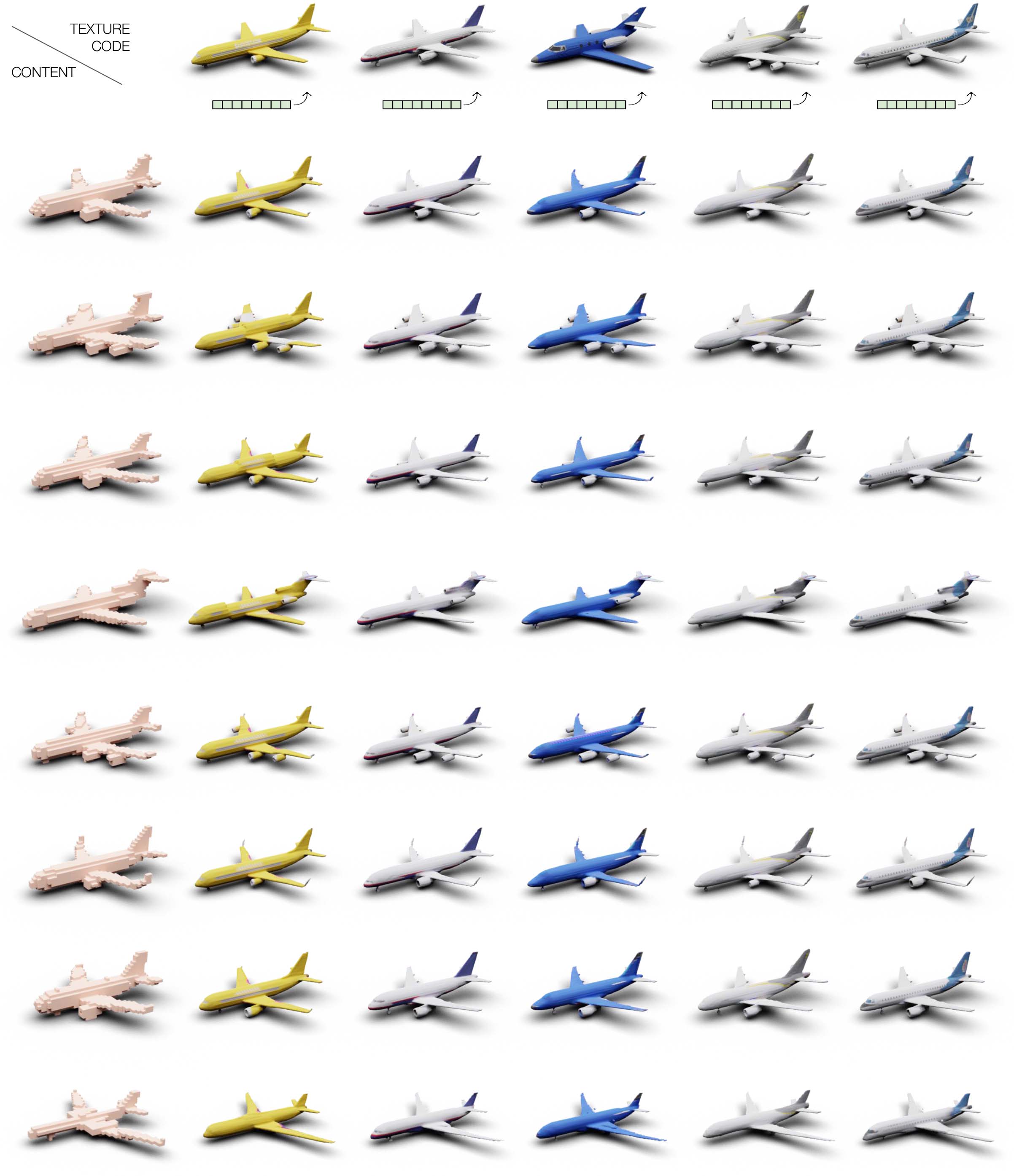}
  \vspace{-4mm}
  \caption{Results of geometry detailization and texture generation on airplane category. Style shapes are shown in the first row and the input coarse voxels are shown in the first column. The input resolution is $64^{3}$ and the output resolution is $512^{3}$.}
  \label{fig:supp_plane_results2}
\end{figure*}

\newpage
\begin{figure*}
  \includegraphics[width=0.94\linewidth]{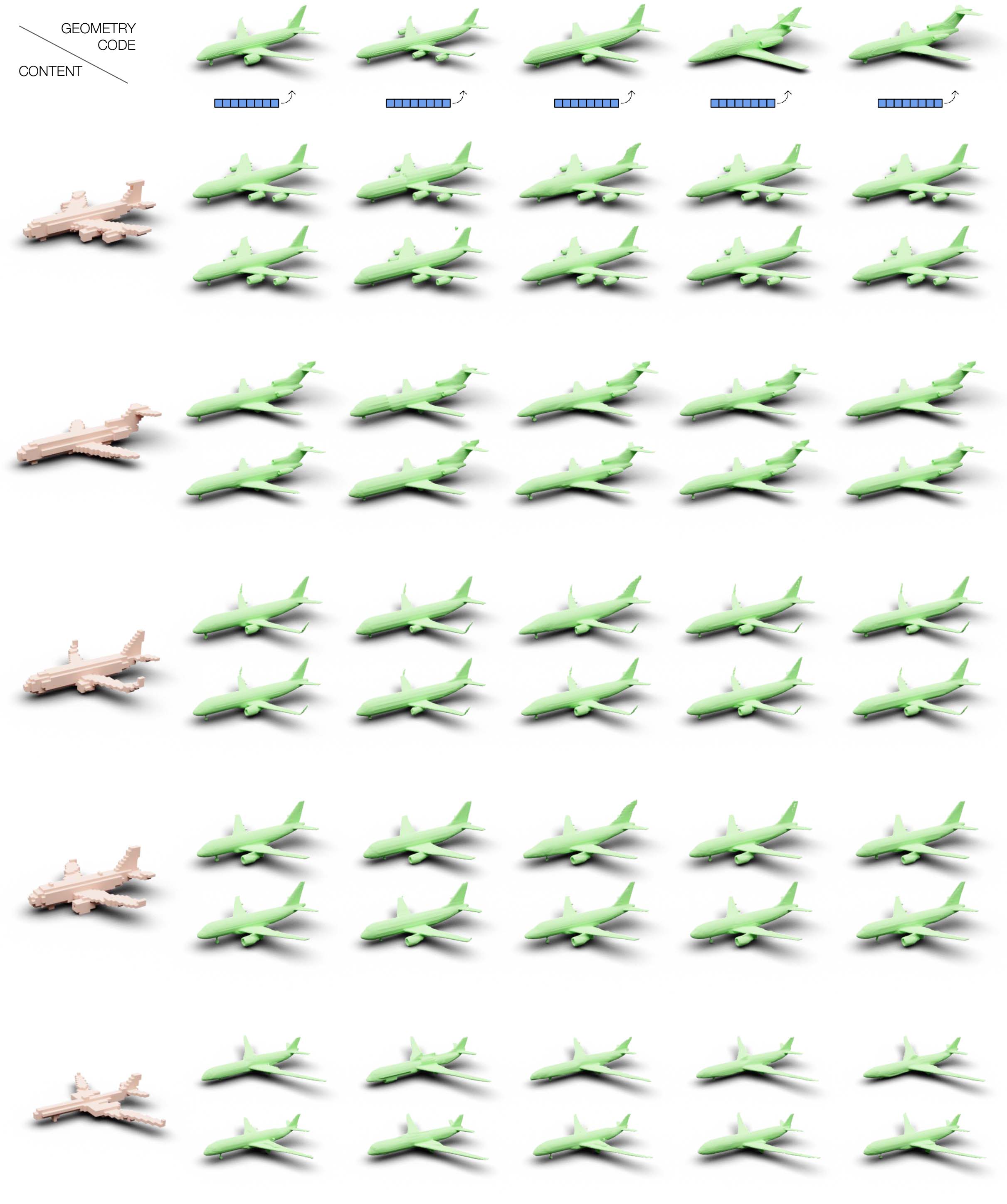}
  \vspace{-2mm}
  \caption{Results of geometry detailization on airplane category. Style geometries are shown in the first row and the input coarse voxels are shown in the first column. Comparisons of DECOR-GAN-up and our multi-resolution are shown in the first and second row of each coarse voxel, respectively. The input resolution is $64^{3}$ and the output resolution is $512^{3}$. Please zoom in to observe the details.}
  \label{fig:supp_plane_results3}
\end{figure*}

\newpage
\begin{figure*}
  \includegraphics[width=0.98\linewidth]{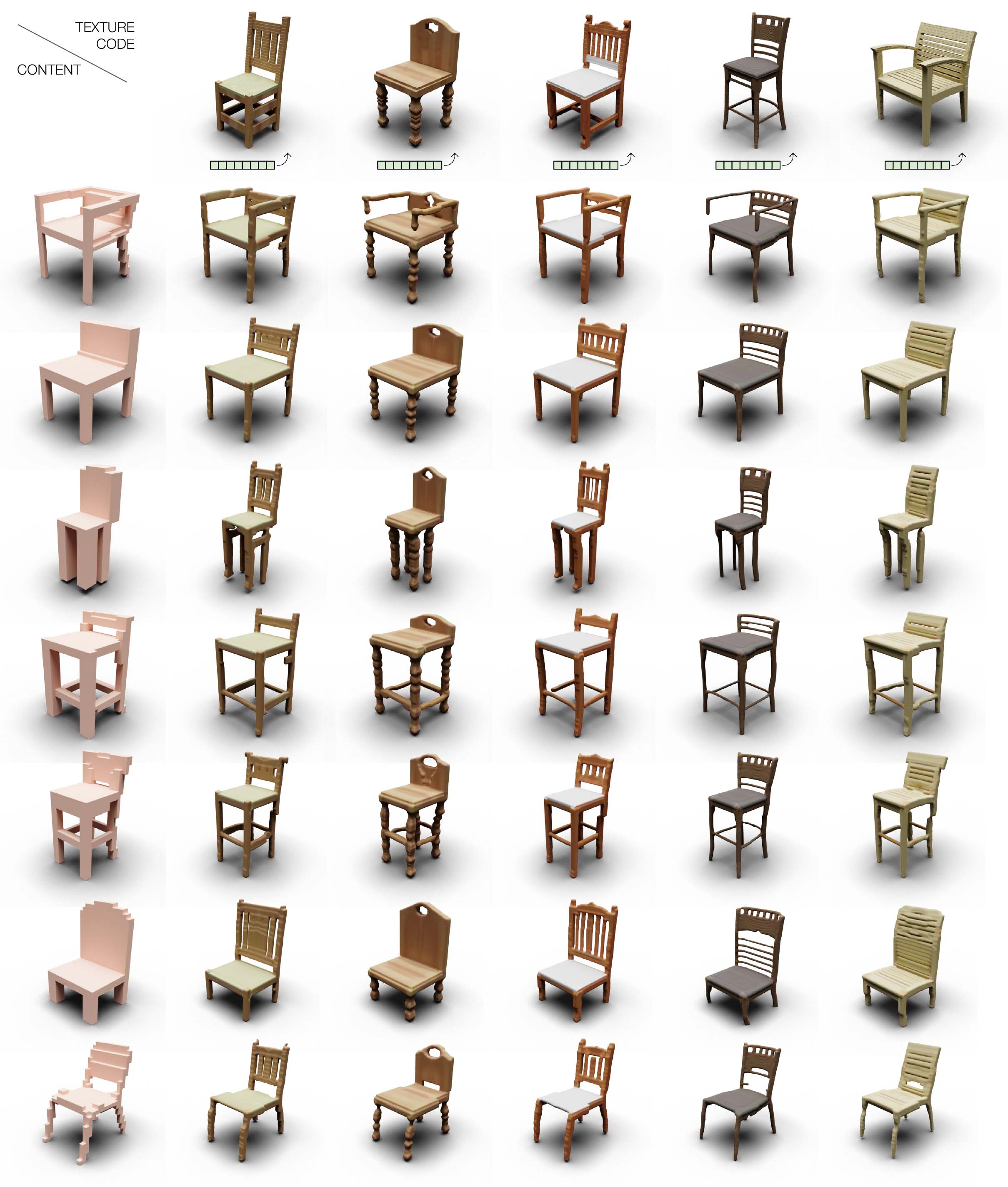}
  \vspace{-4mm}
  \caption{Results of geometry detailization and texture generation on chair category. Style shapes are shown in the first row and the input coarse voxels are shown in the first column. The input resolution is $32^{3}$ and the output resolution is $256^{3}$.}
  \label{fig:supp_chair_results1}
\end{figure*}

\newpage
\begin{figure*}
  \includegraphics[width=0.98\linewidth]{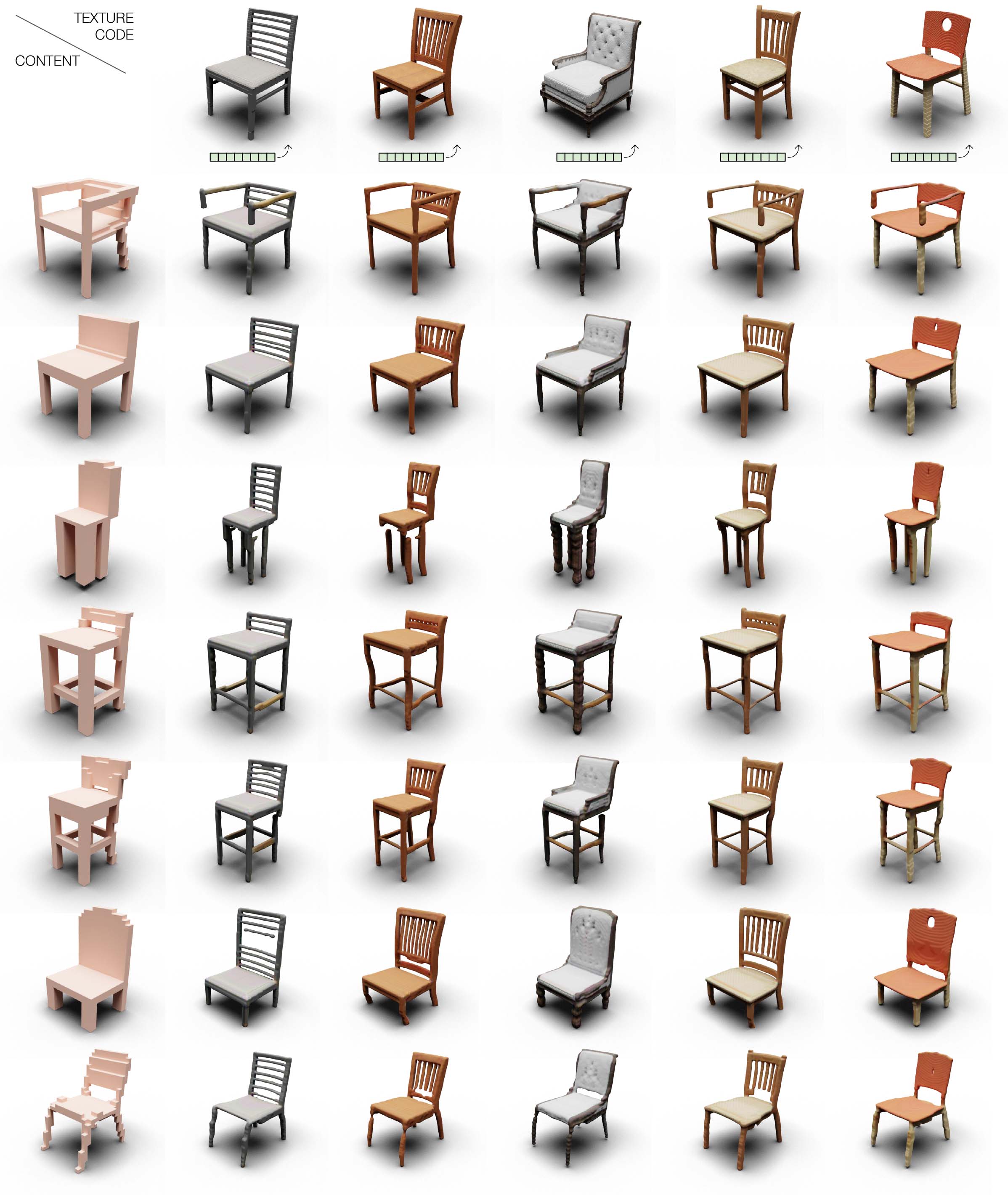}
  \vspace{-4mm}
  \caption{Results of geometry detailization and texture generation on chair category. Style shapes are shown in the first row and the input coarse voxels are shown in the first column. The input resolution is $32^{3}$ and the output resolution is $256^{3}$.}
  \label{fig:supp_chair_results2}
\end{figure*}

\newpage
\begin{figure*}
  \includegraphics[width=0.85\linewidth]{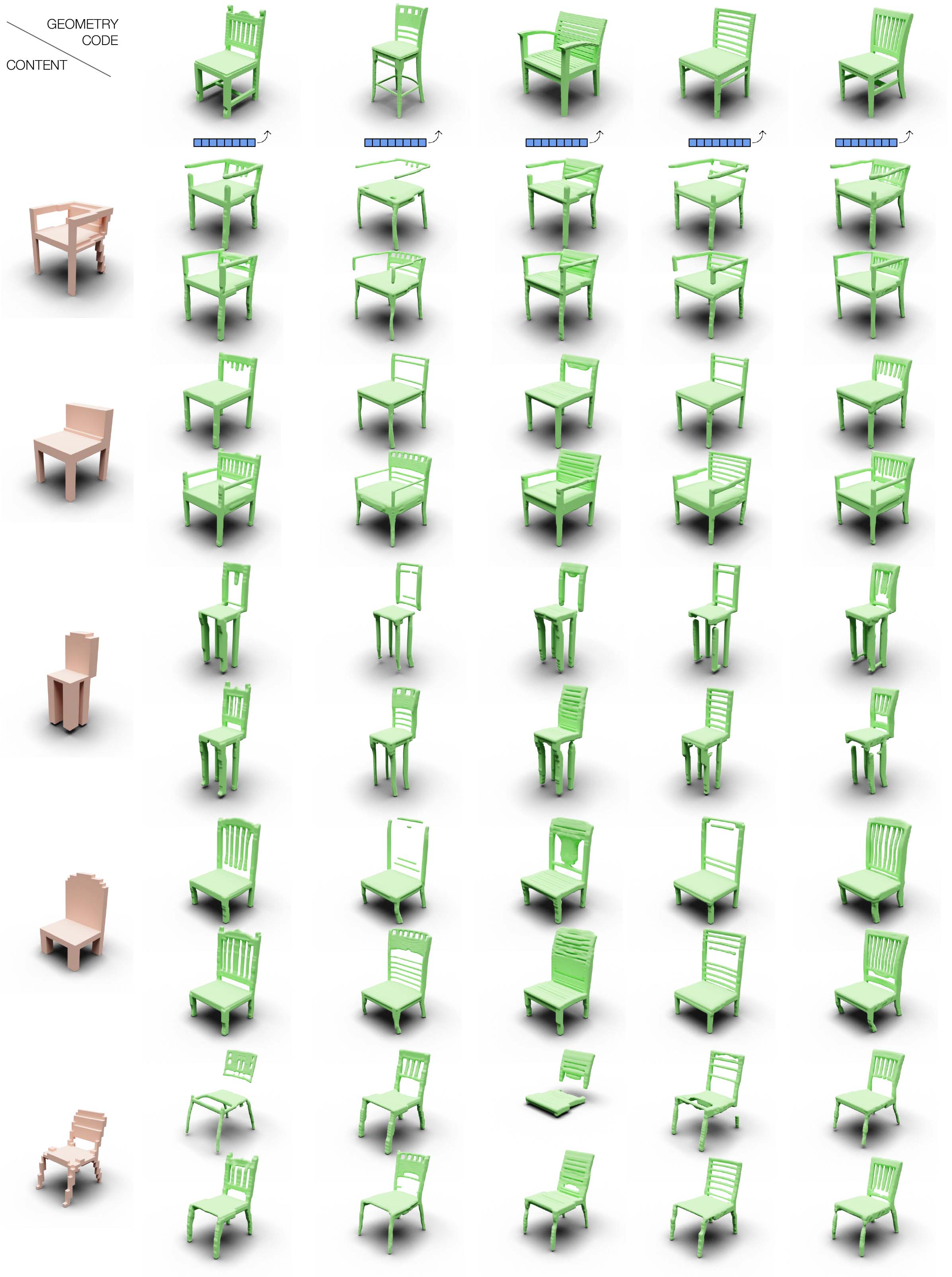}
  \vspace{-2mm}
  \caption{Results of geometry detailization on chair category. Style geometries are shown in the first row and the input coarse voxels are shown in the first column. Comparisons of DECOR-GAN-up and our multi-resolution are shown in the first and second row of each coarse voxel, respectively. The input resolution is $32^{3}$ and the output resolution is $256^{3}$. Please zoom in to observe the details.}
  \label{fig:supp_chair_results3}
\end{figure*}

\newpage
\begin{figure*}
  \includegraphics[width=0.98\linewidth]{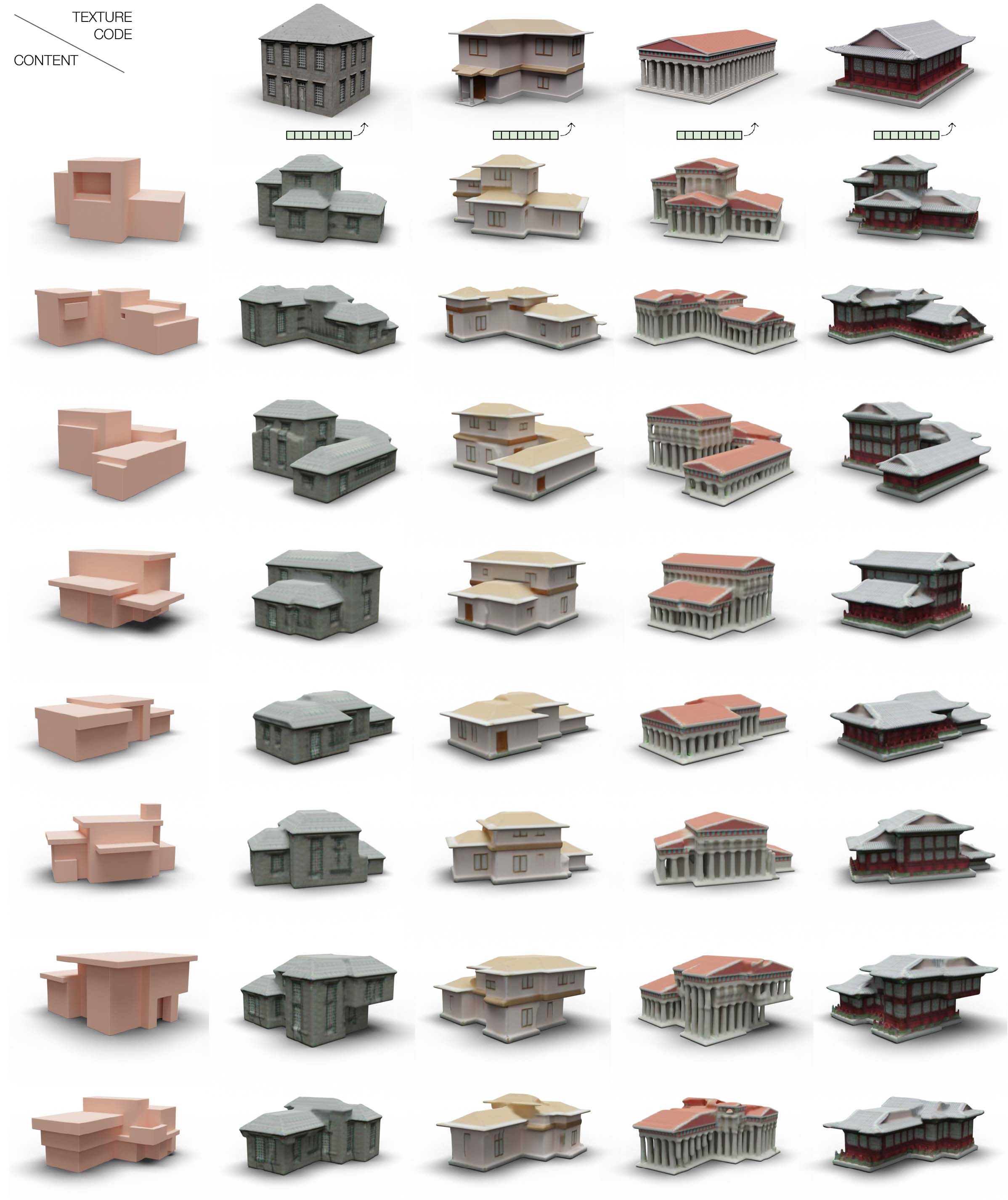}
  \vspace{-4mm}
  \caption{Results of geometry detailization and texture generation on building category. Style shapes are shown in the first row and the input coarse voxels are shown in the first column. The input resolution is $32^{3}$ and the output resolution is $256^{3}$.}
  \label{fig:supp_building_results1}
\end{figure*}

\newpage
\begin{figure*}
  \includegraphics[width=0.98\linewidth]{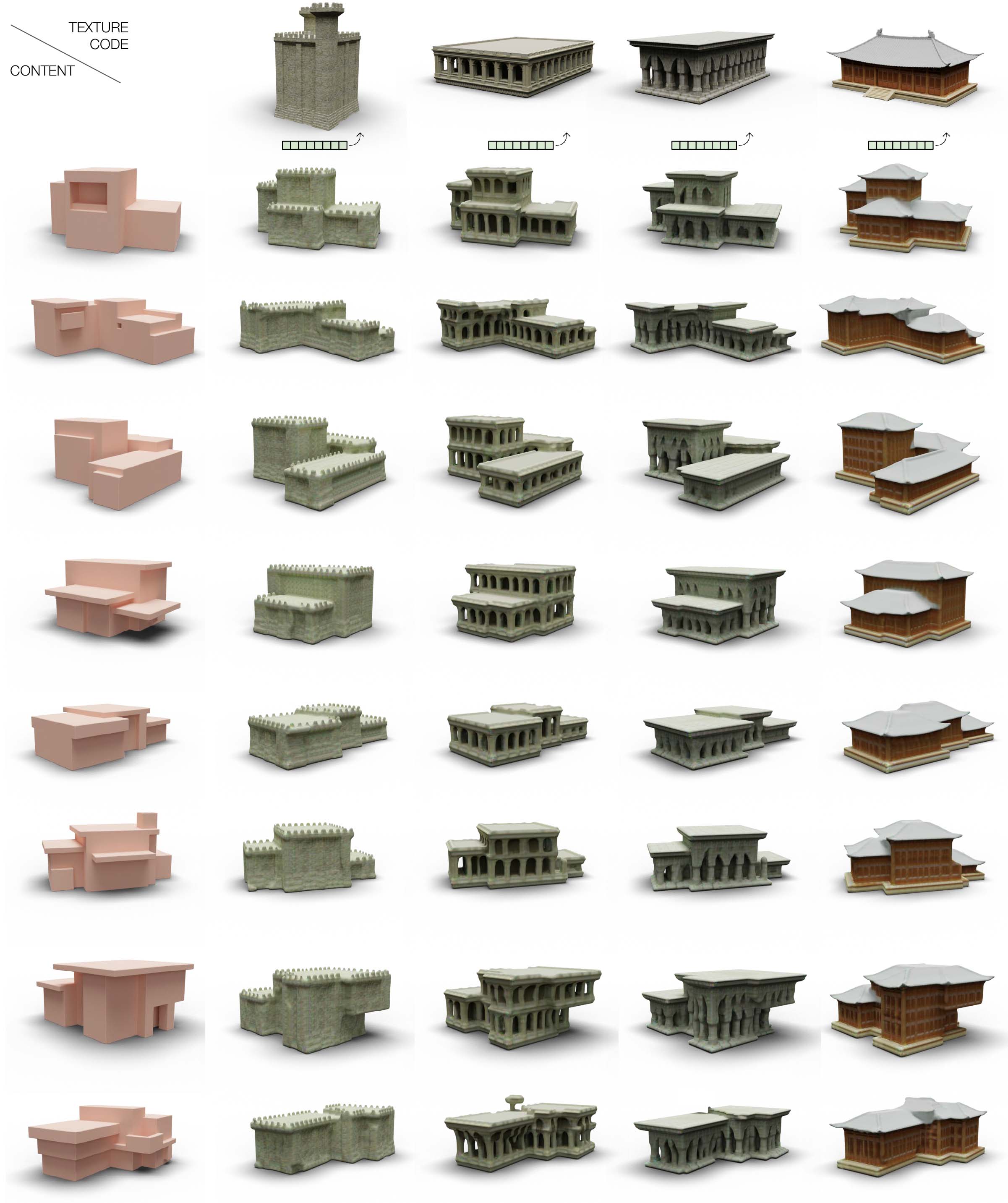}
  \vspace{-4mm}
  \caption{Results of geometry detailization and texture generation on building category. Style shapes are shown in the first row and the input coarse voxels are shown in the first column. The input resolution is $32^{3}$ and the output resolution is $256^{3}$.}
  \label{fig:supp_building_results2}
\end{figure*}

\newpage
\begin{figure*}
  \includegraphics[width=0.9\linewidth]{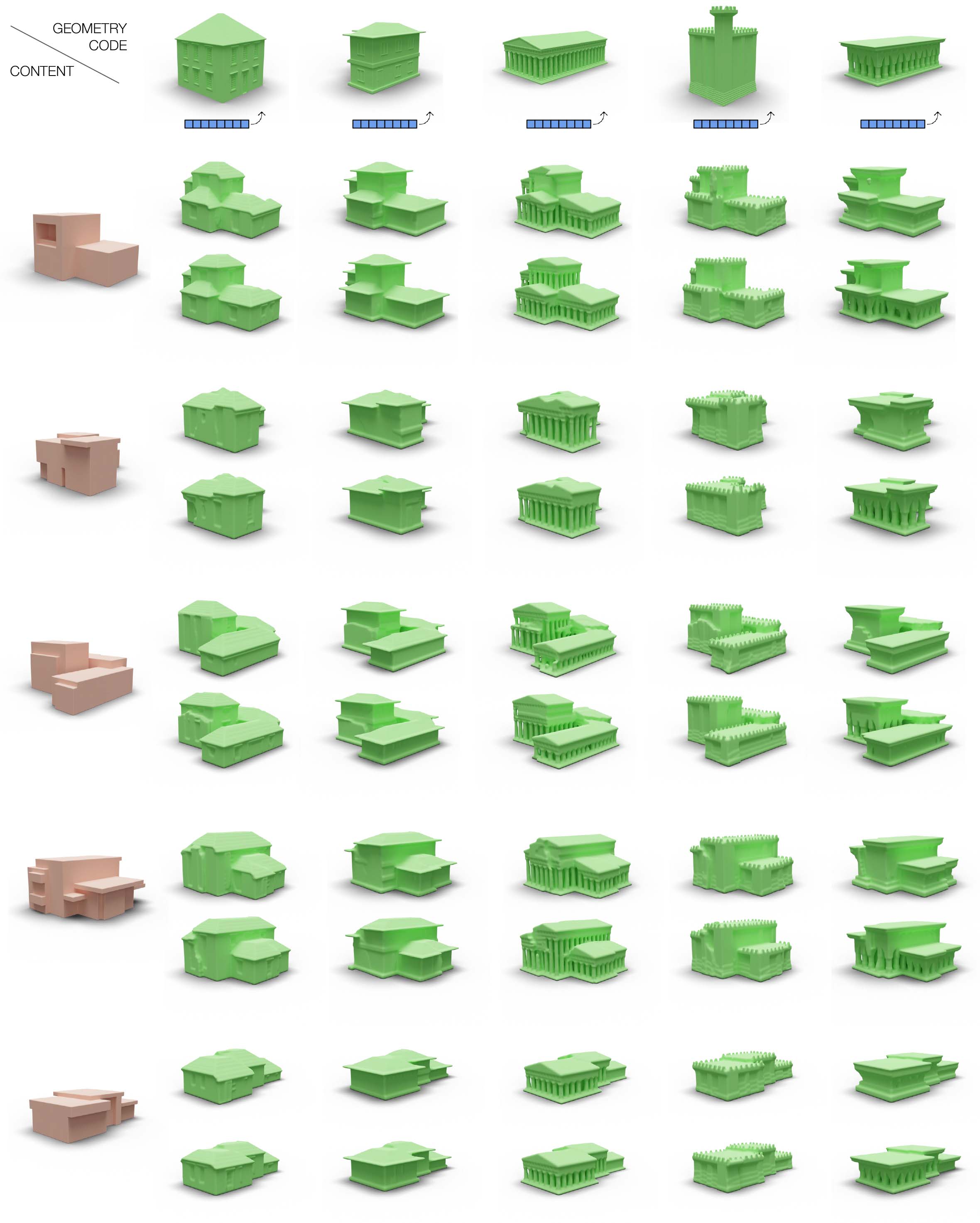}
  \vspace{-2mm}
  \caption{Results of geometry detailization on building category. Style geometries are shown in the first row and the input coarse voxels are shown in the first column. Comparisons of DECOR-GAN-up and our multi-resolution are shown in the first and second row of each coarse voxel, respectively. The input resolution is $32^{3}$ and the output resolution is $256^{3}$. Please zoom in to observe the details.}
  \label{fig:supp_building_results3}
\end{figure*}

\end{document}